\documentclass[screen]{acmart}

\AtBeginDocument{%
  }

\setcopyright{acmlicensed}
\copyrightyear{2025}
\acmYear{2025}
\acmDOI{XXXXXXX.XXXXXXX}

\acmSubmissionID{CSUR-2025-0419}


\usepackage{natbib}
\usepackage[inkscapelatex=false]{svg}
\usepackage{svg}
\usepackage{amsmath}
\usepackage{booktabs}
\usepackage{caption}
\usepackage{subcaption}
\usepackage[normalem]{ulem} 
\usepackage{pdflscape} 
\usepackage{soul}
\usepackage{wrapfig}
\usepackage{float}
\usepackage{adjustbox}

\usepackage{tcolorbox}
\tcbuselibrary{minted,breakable,xparse,skins}
\definecolor{bg}{gray}{0.95}
\DeclareTCBListing{mintedbox}{O{}m!O{}}{%
  breakable=true,
  listing engine=minted,
  listing only,
  minted language=#2,
  minted style=default,
  minted options={%
    linenos,
    gobble=0,
    breaklines=true,
    breakafter=,,
    fontsize=\small,
    numbersep=8pt,
    #1},
  boxsep=0pt,
  left skip=0pt,
  right skip=0pt,
  left=25pt,
  right=0pt,
  top=3pt,
  bottom=3pt,
  arc=5pt,
  leftrule=0pt,
  rightrule=0pt,
  bottomrule=2pt,
  toprule=2pt,
  colback=bg,
  colframe=orange!70,
  enhanced,
  overlay={%
    \begin{tcbclipinterior}
    \fill[orange!20!white] (frame.south west) rectangle ([xshift=20pt]frame.north west);
    \end{tcbclipinterior}},
  #3}

\begin{document}

\title{A Survey on Stereotype Detection in Natural Language Processing}
\author{Alessandra Teresa Cignarella}
\email{alessandrateresa.cignarella@ugent.be}
\orcid{0000-0002-4409-6679}
\affiliation{%
\institution{Language and Translation Technology Team, Ghent University}
\city{Ghent}
\country{Belgium}
}

\author{Anastasia Giachanou}
\email{a.giachanou@uu.nl}
\orcid{0000-0002-7601-8667}
\affiliation{%
\institution{Department of Methodology and Statistics, Utrecht University}
\city{Utrecht}
\country{The Netherlands}}

\author{Els Lefever}
\email{els.lefever@ugent.be}
\orcid{0000-0002-7755-0591}
\affiliation{%
\institution{Language and Translation Technology Team, Ghent University}
\city{Ghent}
\country{Belgium}
}

\renewcommand{\shortauthors}{Cignarella et al.}

\begin{abstract}
\textbf{Abstract.}
Stereotypes influence social perceptions and can escalate into discrimination and violence. While NLP research has extensively addressed gender bias and hate speech, stereotype detection remains an emerging field with significant societal implications. This work presents a survey of existing research, drawing on definitions from psychology, sociology, and philosophy. A semi-automatic literature review was conducted using Semantic Scholar, through which over 6,000 papers (published between 2000–2025) were retrieved and filtered. The analysis identifies key trends, methodologies, challenges and future directions. The findings emphasize the potential of stereotype detection as an early-monitoring tool to prevent bias escalation and the rise of hate speech. The conclusions call for a broader, multilingual, and intersectional approach in NLP studies.
\end{abstract}



\maketitle

\noindent {\Large 
\hspace{1.4cm} \textit{\textbf{Warning:} This paper contains examples of stereotypical and offensive content.}}

\section{Introduction}
The rise of digital environments, like social media, offers great opportunities for sharing ideas, growing businesses, or building communities on a global scale. At the same time, however, this low-barrier and real-time accessibility also fosters the rapid proliferation of harmful content and hate speech. As the boundaries between online and offline spaces increasingly blur, real-world events frequently trigger waves of hate targeting specific demographic groups – whether based on race, religion, gender or sexual orientation – across both spaces. As a very relevant and impactful real-world example, we can refer to the attacks and military response in Israel and Gaza, which have precipitated an alarming rise in threats and violence against both the Jewish and the Muslim community in online and offline settings across Europe. Indeed, reports of the European Commission\footnote{\url{https://www.europarl.europa.eu/RegData/etudes/BRIE/2024/762389/EPRS\_BRI(2024)762389\_EN.pdf}.} signal a considerable increase in hate speech and hate crime over the last twenty years towards both Jews and Muslims across the EU. Expressing hate seems to have become socially acceptable \cite{bilewicz2020hate}, stigmatizing and dehumanizing individuals and groups of people, based on race, age, ethnicity, religion, gender, or sexual orientation, leading to polarization, radicalization and violence. While victims often suffer immediate emotional distress and fear, the longer-term impact can be just as damaging, as members of minority communities adapt their behavior to avoid hate, limiting their full participation or engagement in society \cite{paterson2019short}.

A crucial, yet often under-examined component of hate speech is its foundation in stereotypes - widely held but oversimplified and generalized beliefs about social groups. Stereotypes are not merely cognitive shortcuts used to simplify the cognitive overload of the real world, but they are deeply ingrained social constructs that influence perception, reinforce inequalities, and shape discriminatory behaviors over time \cite{augoustinos1998construction}. Repeated exposure to stereotypes in media, conversations, and online discourse eventually leads to their normalization, making them more socially acceptable and even implicit in everyday interactions. Over time, these internalized biases evolve into prejudices and discriminatory behaviors, fueling societal division and violence. Consequently, hate speech does not arise in isolation: it is the product of entrenched stereotypes that have been reinforced through cultural, political, and technological means. In particular, both hate speech and stereotypes are products of ``shared cultural knowledge, beliefs, and group ideals that create ethnic identities and influence group actions''~\cite{ayansola2021nigeria}, and recent studies have highlighted their intricate relationship~\cite{davani2023socialstereotypes,vargas2023sociallyresponsible}.

Sociological research has long established that stereotypes, prejudice, and discrimination are interconnected, yet distinct phenomena. Although all rooted in bias against the \emph{outgroup}, i.e.,~people outside the own social group or \emph{ingroup}~\cite{dovidio2010prejudice}, a distinction is made in sociology between (1) \emph{prejudice} or emotional bias, when people show negative attitudes or emotions towards the outgroup, (2) \emph{stereotypes} or cognitive bias, which are based on beliefs or cognitive associations about group characteristics, and (3) behavior bias expressed through \emph{discrimination}, which involves actions or practices that disadvantage certain demographic groups~\cite{fiske1998stereotyping,stangor2000stereotypes}. Although considered to be different types of bias, researchers have argued that stereotypes and prejudices are linked, with prejudice often stemming from stereotypic beliefs~\cite{fiske1998stereotyping,dovidio2010prejudice}.

In our research, we focus on stereotypes, understood as category-based generalizations that not only reflect ``beliefs about the traits characterizing typical group members but also contain information about other qualities such as social roles and the degree to which members of the group share specific qualities''~\cite{dovidio2010prejudice}. 
Understanding and detecting stereotypes is particularly important because they operate not only at the explicit level but also implicitly, shaping thought processes and behaviors outside of conscious awareness. Even individuals who consciously reject stereotypes can still exhibit implicit biases \cite{greenwald1995implicit}. These implicit biases are molded by social and cultural experiences and can impact decision-making in areas such as hiring, policing, and healthcare. For example, ~\citet{hoffman2016racial} showed that medical professionals may unintentionally provide different levels of care to patients based on their race, often offering less pain relief to black patients compared to white patients, despite similar reported pain levels.\footnote{Based on the stereotypical assumption that black people experience less pain because they have thicker skin or less sensitive nerve endings).} 

A well-known method to measure implicit bias is the Implicit Association Test (IAT), a widely used tool in social cognition research to measure associations between concepts and attributes~\cite{greenwald1998measuring}. While mainly used to detect implicit attitudes, recent research shows its output may also represent explicit attitudes~\cite{tahamata2024iat}. The test involves categorizing stimuli into concept and attribute pairs, with faster responses indicating stronger associations~\cite{houwer2001structural}. The tool claims to also reveal implicit attitudes of which the individuals taking the test were unaware.\footnote{ \url{https://implicit.harvard.edu/implicit/takeatest.html}.}
While the IAT measures implicit biases which are influenced by broader societal stereotypes, the Stereotype Content Model (SCM) explains how these stereotypes are formed. SCM stems from social psychology and proposes to classify stereotypes along two dimensions or axes: \emph{warmth}, or encompassing sociability and morality -- which is associated with cooperative groups, and \emph{competence}, or encompassing ability and agency -- which is associated with high-status groups~\cite{fiske2002modelmixed}. On one end of the spectrum, the ingroup is often perceived as both warm and competent, while on the opposite end are social outcasts, such as homeless individuals, who are frequently stereotyped as not being cooperative and being incompetent. Certain demographic groups may score high on one dimension but low on the other, such as older people that are often viewed as high in warmth but low in competence.\footnote{Example from Fiske: \url{https://nobaproject.com/modules/prejudice-discrimination-and-stereotyping}.}

To acknowledge the role of subjectivity, the SCM model was later expanded by \citet{koch2016abc}, culminating in the Agency-Beliefs-Communion (ABC) model.
This newer model directly stems from the SCM but introduces a crucial third dimension: \textit{beliefs}. While the SCM focuses on how people stereotype groups based on warmth and competence, the ABC model also explains how people form impressions of individuals by considering whether they share our same values and beliefs. 

Imagine you meet a new coworker. The SCM model \cite{fiske2002modelmixed} would suggest that you judge them based on \textit{warmth} (Are they friendly?) and \textit{competence} (Are they good at their job?). If they are both friendly and skilled, you admire them. If they seem unfriendly but competent, you may resent them. On the other hand, the ABC model \cite{koch2016abc,koch2020groupswarmth} suggests that beyond warmth and competence, you also consider their \textit{beliefs} , i.e., do they share your values or opinions? For example, if your coworker is friendly and competent but has completely different political or ethical views, you might still feel distant from them.

With this paper, we aim to provide a comprehensive survey of research addressing different aspects of stereotype detection. Although our main focus is on studies that examine stereotypes from the NLP aspect, we also include theoretical contributions that cover a sociological perspective of the problem. Stereotype detection is a rather new field in NLP and computational linguistics, and as explained before, is closely related to the fields of bias and hate speech detection. In particular, although gender bias has received substantial scholarly attention, most of this work has focused on English and other high-resource languages.

Other limitations for the detection and mitigation of gender bias are the strict definition of gender as a binary variable, and lacking evaluation baselines and pipelines~\cite{stanczak2021surveygenderbias}.
This survey is inspired by the survey on automatic detection of hate speech by \citet{fortuna2018survey}, who provide an overview of the research on hate speech detection; its definition, applied methods and resources.

\begin{figure}
\includegraphics[width=0.77\textwidth]{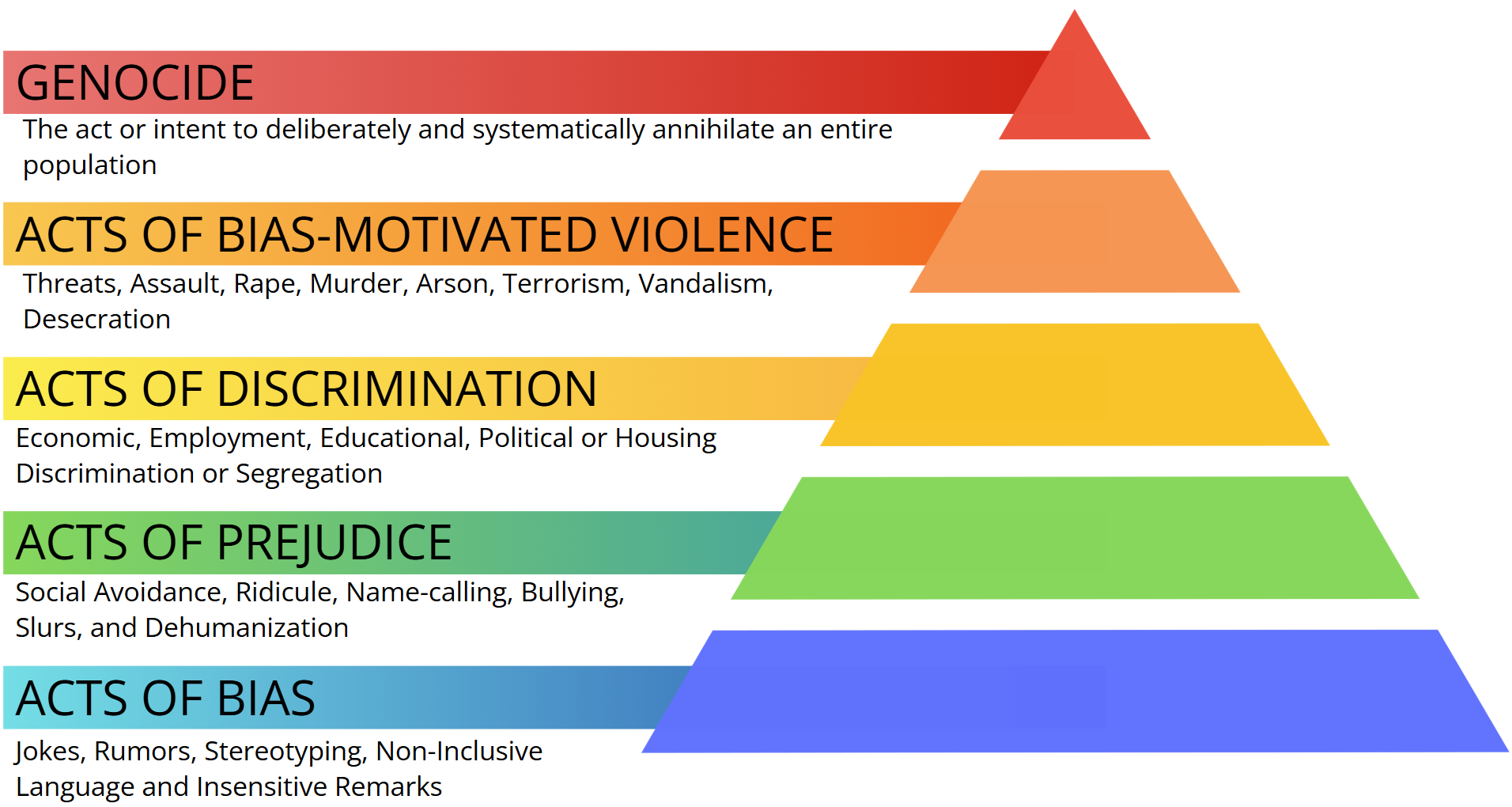}
\caption{The Pyramid of Hate.}\label{fig:pyramid}\Description{}
\end{figure}

We believe stereotype detection is an important NLP task with potentially a far-reaching societal impact, as it allows bias to be detected in its early-stage manifesting, before it turns into hate or violence and it can, as such, be helpful to counteract them. This is inspired by the \emph{Pyramid of Hate}, illustrated by Figure~\ref{fig:pyramid}, that shows how killing and other acts of violence might often be the extreme consequences that dangerously evolve from too-often overlooked acts of bias and prejudice, such as offensive jokes, offensive humor, non-inclusive language, and stereotypes.\footnote{Adapted from the \href{https://www.adl.org/sites/default/files/pyramid-of-hate-web-english_1.pdf}{Anti-Defamation League}.}

Finally, we should note that stereotype detection in NLP is closely connected to adjacent areas such as hate speech detection, bias evaluation in language models, and misinformation analysis, areas that have received extensive research interest. Several researchers have shown that stereotypes and hate speech are correlated \cite{liu2024quantifying}. Similar to hate speech detection, stereotype identification focuses on harmful language patterns, but often targets more subtle or implicit expressions that may not meet the threshold of overt toxicity, yet still perpetuate social harms \cite{davani2023socialstereotypes}. Similarly, studies on bias in language models have demonstrated how stereotypes are encoded and amplified by large-scale pre-trained models, with biases such as dialect prejudice or gendered associations persisting even after mitigation efforts \cite{hofmann2024ai}. Stereotypes can also intersect with misinformation, where biased narratives are used to distort facts or reinforce harmful group generalizations. The relevance of stereotype detection to those areas underscores its importance toward building fairer, and more reliable NLP systems.

\subsection{What are Stereotypes?}

\noindent As mentioned earlier, stereotypes are widely held, simplified, and often generalized beliefs about a particular group of people.
\citet{fiske2000stereotyping} defines stereotypes as cognitive structures that contain knowledge, beliefs, and expectations about social groups. These mental shortcuts help individuals navigate social interactions but can also reinforce biases and perpetuate discrimination.  
Beyond psychology, sociology offers a broader perspective on stereotypes, viewing them as socially constructed narratives that influence group identities, power dynamics, and social inequalities. Sociologist \citet{lippmann1922stereotypes} was one of the first to conceptualize stereotypes as ``pictures in our heads'', highlighting how they shape perceptions and maintain social hierarchies. Meanwhile, in philosophy, scholars like \citet{haslanger2000feminism} argue that stereotypes play a role in sustaining oppressive structures, particularly in relation to race, gender, and class.  

Institutional frameworks also address stereotypes, particularly in relation to gender. The \href{https://end-gender-stereotypes.campaign.europa.eu/index_it}{\ul{European Union}} and the \href{https://www.ohchr.org/en/women/gender-stereotyping}{\ul{United Nations}} define gender stereotypes as preconceived ideas about the roles and characteristics of women and men, often leading to unequal treatment and reinforcing structural barriers. While gender is a significant dimension, it is essential to broaden the scope of stereotype research to include ethnicity, disability, sexual orientation, and other social categories. A more comprehensive approach allows for an intersectional analysis, recognizing how multiple identity factors interact to shape experiences of bias and discrimination. Intersectionality, a concept introduced by \cite{crenshaw1989intersectionality}, highlights how overlapping social identities create unique experiences of oppression, making it crucial to study stereotypes beyond isolated categories. By incorporating an intersectional perspective, we can better understand how different forms of bias intersect and reinforce one another in social and linguistic contexts.

In this work, we define stereotypes as recurrent, socially constructed (generalized) associations that ascribe fixed characteristics to individuals based on their group membership. These associations often simplify complex realities, reinforcing bias and prejudices and limiting freedom and opportunities for those subjected to them. Stereotypical text can manifest in various ways across different target groups.\\ 
For example:
\begin{itemize}
\item \ul{Gender stereotypes}: ``Women are naturally bad at driving'' or ``Men don’t know how to show emotions'';
\item \ul{Ethnic stereotypes}: ``Asian people are good at math'' or ``Latinos are always passionate and loud'';
\item \ul{Age stereotypes}: ``Older employees struggle with technology'' or ``Young people are lazy and entitled'';
\item \ul{Disability stereotypes}: ``People with disabilities are always inspirational'' or ``People with autism lack empathy'';
\item \ul{LGBTQIA+ stereotypes}: ``Gay men are always flamboyant'' or ``Lesbians are just tomboys who hate men''.
\end{itemize}

\noindent These examples illustrate how stereotypes reduce individuals to predefined categories, influencing social perception and reinforcing systemic biases. Addressing and analyzing such stereotypes is crucial for developing fair and inclusive NLP models capable of mitigating their harmful effects.

\subsection{Scope of this Survey}
This review paper aims to provide an overview of stereotype detection in NLP, summarizing and discussing theoretical frameworks stemming from psychology studies, existing methodologies, datasets, challenges, and future directions. While previous research has extensively focused on hate speech and mitigating stereotypical bias in AI systems, stereotypes, which are a crucial aspect of harmful language, have only recently attracted research attention. 

Our focus on stereotypes, rather than hate speech, is motivated by their subtle, yet pervasive, influence on language and social interactions. Stereotypes serve as the cognitive foundation for prejudice and discrimination, shaping societal attitudes and influencing behaviors both explicitly and implicitly. Hate speech is often explicit, characterized by hostility and derogatory expressions targeting specific groups. Stereotypes, in contrast, can manifest in more implicit and normalized ways, reinforcing social hierarchies and biases without overtly aggressive language. This distinction is important because harmful stereotypes can propagate discrimination even in the absence of explicit hate speech, making their detection vital for inclusiveness-aware AI development. Early identification of stereotypes can serve as a preventative measure, reducing the likelihood of their escalation into more explicit manifestations of bias and discrimination.

Please note that this survey is \textit{not} a systematic review, since conducting one would require an exhaustive analysis of all 139+ papers related to stereotype detection in NLP that were retrieved (see details Section \ref{sec:methodology}). Instead, we tried to find patterns based on the retrieved papers to provide a broad yet selective overview of relevant work, focusing on current key trends, methodologies, and gaps in the field. Ultimately, this survey seeks to be a starting point for scholars who are now entering this research topic, offering a structured understanding of what has been done so far and identifying open challenges. A more exhaustive discussion is provided in the \hyperlink{h:limitations}{Limitations} section at the end of this paper.

Compared to existing surveys, our work offers a distinct and timely contribution by centering specifically on stereotype detection in NLP, a topic that has often been treated as a subcomponent of broader bias or hate speech research. {While} \citet{fortuna2018survey} and \citet{poletto2021resources} {provide foundational overviews of hate speech detection and its resources, they do not isolate stereotypes as a unique linguistic phenomenon.} \citet{stanczak2021surveygenderbias} and \cite{bartl2025gender} {focus on gender bias, with the latter extending the discussion to computer vision, but both remain limited to specific bias types and modalities.} \citet{garg2023handling} {address bias in toxic speech detection, emphasizing fairness and robustness, yet their scope is confined to overtly harmful content. In contrast, our survey highlights the subtle, implicit, and often socially normalized nature of stereotypes, which may not trigger toxicity filters but still perpetuate harm. We also uniquely map the connections between stereotype detection and adjacent areas such as hate speech, bias in language models, and misinformation, offering a cross-cutting perspective that underscores the relevance of stereotype analysis for building fairer NLP systems. Furthermore, by systematically categorizing stereotype types, detection methodologies, and evaluation strategies, this survey offers a structured and comprehensive resource that addresses a notable gap in the existing literature. For a summarized comparison with related surveys, see} \autoref{tab:survey_comparison}.

\begin{table}[!ht]
\centering
\small
\begin{tabular}{p{3.5cm}p{4.7cm}p{6cm}}
\hline
\textbf{Survey Paper} & \textbf{Focus Area} & \textbf{Scope and Key Contributions} \\
\hline
\citet{fortuna2018survey} & Hate speech and bias in NLP & Surveys hate speech detection and bias-related NLP tasks; reviews definitions, datasets, and detection methods; emphasizes challenges in generalization and the overlap with stereotypes. \\
\hline
\citet{stanczak2021surveygenderbias} & Gender bias in NLP & Reviews 304 papers focused on gender bias; analyzes definitions of gender, surveys datasets and mitigation methods; critiques binary framing and highlights lack of multilingual and ethical considerations. \\
\hline
\citet{poletto2021resources} & Hate speech detection (resources) & Systematic review of benchmark corpora and lexica; analyzes development methodologies, topical focus, and language coverage; identifies gaps and opportunities for improvement. \\
\hline
\citet{garg2023handling} & Bias in toxic speech detection & Reviews bias mitigation in toxicity detection; examines evaluation methods; introduces concept of bias shift; highlights fairness and robustness challenges in toxicity classifiers. \\
\hline
\citet{bartl2025gender} & Gender bias in NLP and computer vision & Comparative survey across textual, visual, and multimodal models; identifies conceptual parallels and methodological transfers between NLP and CV; emphasizes interdisciplinary approaches and critiques binary gender framing. \\
\hline
\textbf{\ul{This Survey}} & \textbf{\ul{Stereotype detection in NLP}} & Broad coverage across stereotype types, datasets, methods, and evaluation; connects stereotype detection to hate speech, bias in LMs, and misinformation; categorizes approaches; highlights subtle harms, implicit bias and the need for intersectional research. \\
\hline
\end{tabular}
\caption{Chronological comparison of survey papers related to bias, stereotypes, and hate speech in NLP and adjacent fields.}
\label{tab:survey_comparison}
\end{table}

\noindent 
The rest of the survey is structured as follows: Section \ref{sec:methodology} outlines the methodology adopted for conducting the literature review.
Section \ref{sec:main-literature-categories} presents an organized overview of the literature, categorizing works based on their focus, techniques, and findings related to stereotypes in NLP. 
Section \ref{sec:challenges} discusses the main challenges faced in this field, including limitations in data, evaluation metrics, and language coverage. 
Finally, Section \ref{sec:conclusion} concludes the survey by summarizing the insights gained and outlining future research directions, with Section \ref{sec:opportunities} highlighting specific opportunities for advancing stereotype detection in NLP.
\section{What NLP research has been conducted so far regarding \emph{Stereotype Detection}?\label{sec:methodology}}

To gain an insight into NLP researchers' perspectives, methodologies and types of research on the topic of \emph{Stereotype Detection}, we conducted an automated search to identify relevant literature. This section outlines the methodology employed and presents a detailed analysis of the results.

\subsection{Methodology of the Literature Review}
\begin{figure}[!h]
\centering
\includegraphics[width=\textwidth]{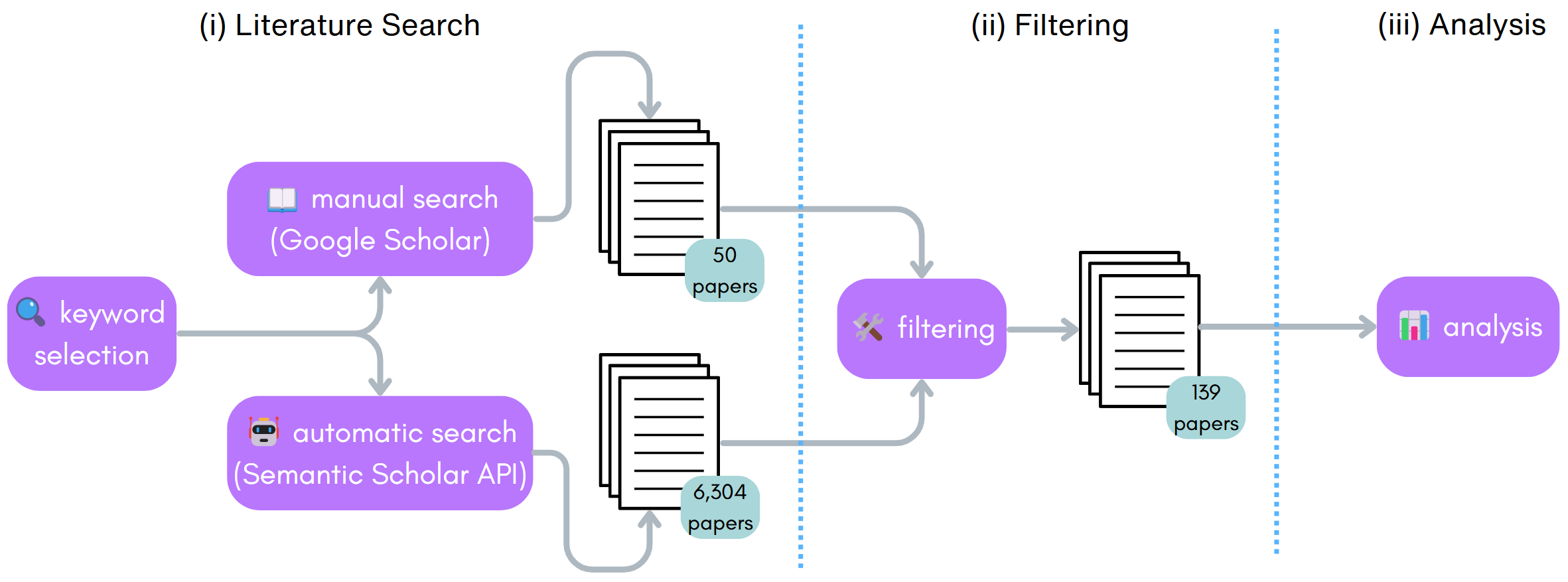}
\caption{Pipeline describing the different steps of the survey process.}\label{fig:image-pipeline-phases}\Description{}
\end{figure}

Our methodology was designed to ensure a comprehensive approach to identifying relevant literature. The process began with selecting appropriate keywords, followed by a manual search to obtain an initial set of \textit{seed} papers. 
Later on, an automatic search was conducted, retrieving up to 1,000 papers per year from 2000 to 2025.
The results of the automatic search were filtered and cross-referenced with previously known works on the topic to ensure completeness. 
We then performed a cross-search to identify additional relevant publications, followed by an automatic filtering process to refine the dataset.
Finally, some bibliometric analyses were conducted to provide insights into the research trends, to highlight the most recurrent patterns, and the key contributions within the body of literature analyzed. The methodology can be divided into three main phases: (i) Literature Search, (ii) Filtering, and (iii) Analysis. In \autoref{fig:image-pipeline-phases}, we outline the steps of the adopted pipeline. In the subsections below, we describe the three phases in detail.

\subsubsection{\textbf{Literature Search}\label{subsubsec:automatic-search}}
In the initial stage we conducted a manual search on Google Scholar\footnote{\url{https://scholar.google.com/}} using the keywords <\textsf{stereotype} + \textsf{detection}> and <\textsf{stereotype} + \textsf{nlp}>. 
To ensure the creation of a relevant database, we limited our selection to the first five pages of results (approximately 50 papers), sorted by relevance and without applying any time filter. We selected only open-access papers.
Despite its popularity among fellow researchers, Google Scholar posed significant challenges due to its lack of an official API.

To overcome these limitations, therefore, we decided to rely on the \textit{Semantic Scholar API}\footnote{\url{https://www.semanticscholar.org/product/api}}, which offers a fairer, more robust and compliant solution for retrieving academic literature. We searched for a maximum of 1,000 articles per year over a 26-year span (2000–2025), using the query <\textsf{stereotype} + \textsf{detection} + \textsf{nlp}> within the \textit{Paper Bulk Search}. For every hit of the search, we gathered the following information: title, publication year, abstract, field of study (e.g. engineering, chemistry, linguistics), type of publication (e.g. review, journal article, conference), publishing venue (e.g. Nature, IEEE, New England Journal of Medicine), and the citation count.\footnote{Please note that the last version of the database was created on February 10, 2025.}

As shown by the varying numbers in \autoref{tab:retrieval-automatic-search}, some entries in the database were incomplete. Consequently, a filtering step was necessary to ensure the data's consistency and reliability (see Section \ref{subsubsec:filtering}).
\begin{table}[!ht]
\centering
\begin{tabular}{lr}
\toprule
Title & 6,304 \\
Year & 6,304 \\
Abstract & 5,762 \\
Field of Study & 4,501 \\
Publication Type & 4,883 \\
Venue & 5,126 \\
Citation Count & 4,365 \\
\bottomrule
\end{tabular}
\caption{Resulting size of the database after the search on Semantic Scholar.\label{tab:retrieval-automatic-search}}
\end{table}
However, before proceeding with more stringent filtering, {we sought to gain an initial overview of the collected documents by identifying key themes that significantly overlapped with the query terms <\textsf{stereotype detection}>} across the corpus of over 6,000 collected papers.  

\begin{figure}[h]
\centering
\adjustbox{trim=80 0 0 0,clip}{\includegraphics[scale=0.42]{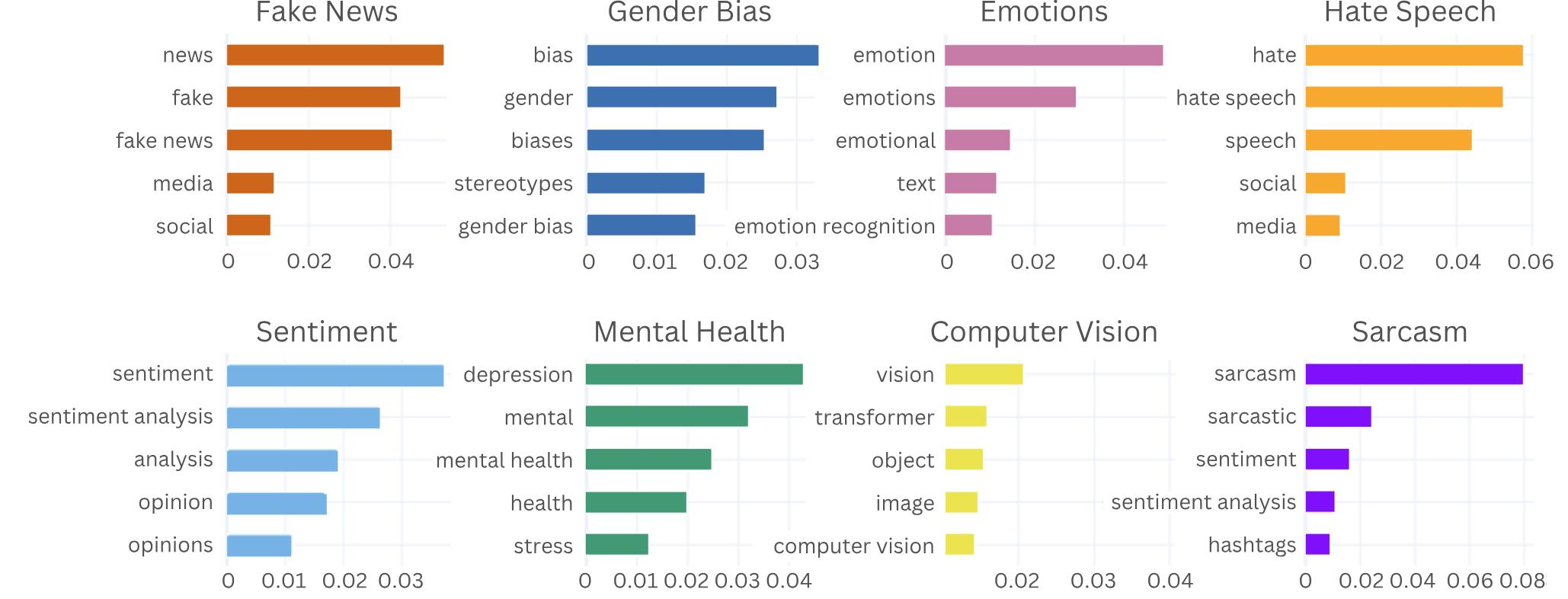}}
\caption{Topics extracted with BERTopic from the text of the abstracts.\label{fig:8-topics}}
\Description{}
\end{figure}

\noindent To achieve this, we applied BERTopic\footnote{\url{https://maartengr.github.io/BERTopic/index.html}} on the 6,304 collected papers to analyze the texts of all abstracts (which were pre-processed by removing punctuation, stop words and query-related terms). 
{We chose this tool because it offers an easy-to-implement and off-the-shelf solution for exploratory topic modeling. Furthermore, it is compatible with widely used visualization libraries such as \textsf{Plotly} and \textsf{matplotlib}.}
The details on the parameters we used with BERTopic can be found in \autoref{appendix}. To ensure relevance was maintained, we set a minimum threshold of 10 papers per topic. 
The model then identified eight key topics closely related to \textit{Stereotype Detection} (in descending order): Fake News, Gender Bias, Emotions, Hate Speech, Sentiment, Mental Health, Computer Vision, and Sarcasm. The relevance of each topic is presented in \autoref{fig:8-topics}.

Despite the valuable insights provided by the topic modeling, the dataset still contained considerable noise. {To address this, we implemented a careful filtering process based on a combination of selected keywords and manual review. This step-by-step refinement was conducted independently of the BERTopic results to ensure the retention of meaningful content.}

\subsubsection{\textbf{Filtering}\label{subsubsec:filtering}}
\begin{figure}[h]
\centering
\includegraphics[width=\linewidth]{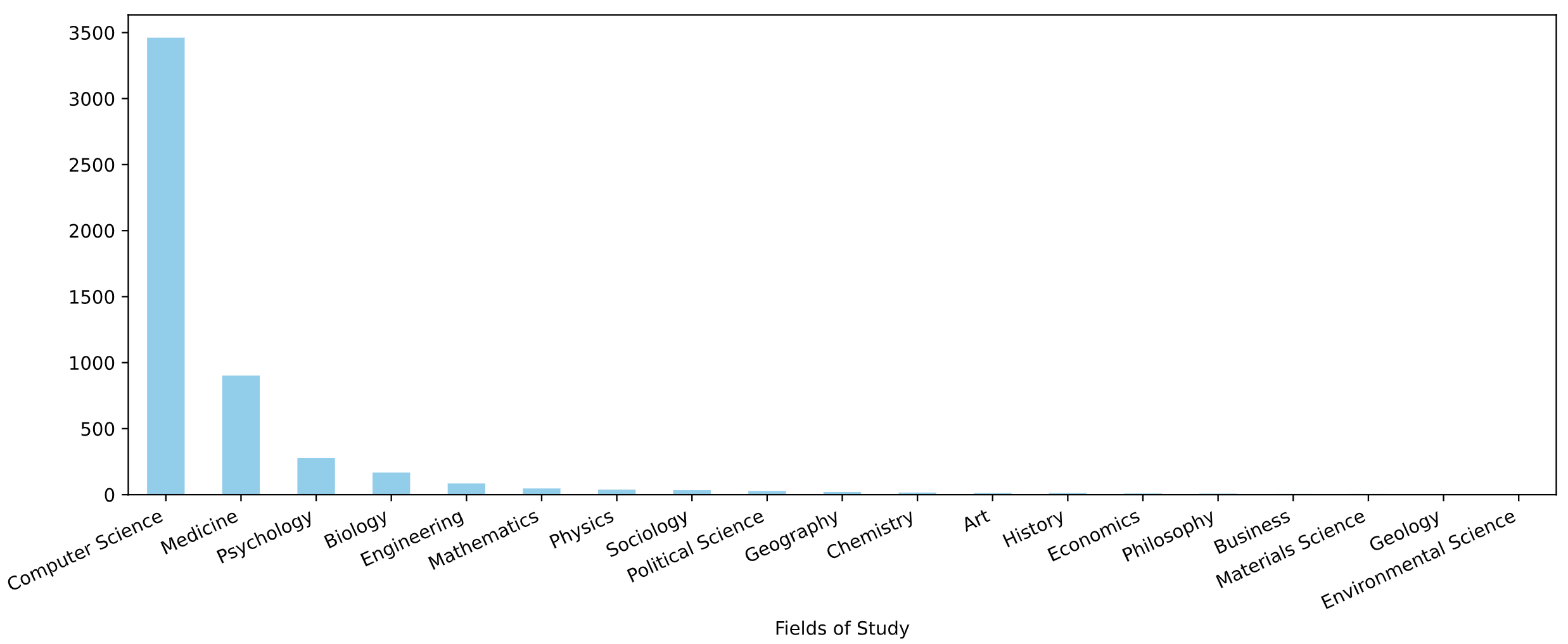}
\caption{Number of publications per field of study.\label{fig:number-by-fieldofstudy}}
\Description{}
\end{figure}

To ensure no relevant data got lost, we approached the filtering process gradually, in small and careful steps.
First, we removed all entries with corrupted data from the database (i.e., if any of the retrieved fields shown in \autoref{tab:retrieval-automatic-search} was empty).
Secondly, we manually inspected some random samples of the database, and observed that despite including the word < \textsf{nlp} > in the original search query on Semantic Scholar, the results were still relatively noisy, containing many entries belonging to disciplines and fields outside the scope of this survey. In \autoref{fig:number-by-fieldofstudy} we display the number of articles per field of study. As is immediately obvious, the main subjects that seem related to the use of the words <\textsf{stereotype detection}> are Computer Science, Medicine, Psychology, Biology and Engineering.

\noindent We performed a semi-automatic review of the downloaded articles, by extracting the most common words (from titles and abstracts) and their relative frequency, and this analysis revealed that a significant number of papers contained terms such as \textsf{autism spectrum}, \textsf{sound event}, \textsf{motor movement}, \textsf{dcase}, \textsf{intrusion detection}, \textsf{bomb attack}, \textsf{child abuse}, \textsf{blue whale}, and \textsf{stimuli} which are not relevant for our goal. 
To address the removal of this extra noise, a more thorough filtering process was performed. 
Therefore, we began by removing papers from scientific areas unrelated to our focus, such as Medicine, Chemistry, Biology, Physics, Geography, Geology, Mathematics, Environmental Science, Engineering, Art, History, Economics, Political Science, Business, and Materials Science. 
After this filtering steps, the number of papers was reduced from 6,304 to 4,959.
In the final filtering step, we retained only those papers where the terms <\textsf{stereotype}> or <\textsf{stereotypes}> or <\textsf{stereotyped}> or <\textsf{stereotyping}> \emph{AND} <\textsf{nlp}> or <\textsf{natural language processing}> appeared in the title \emph{OR} in the abstract. This ensured the dataset remained focused on the topic of \textit{Stereotype Detection in NLP}. The final count amounts to \textbf{\ul{139 papers}}.

\subsubsection{\textbf{Analyses}\label{subsubsec:analyses-on120paers}}
The analyses that follow are thus carried out on the final database, consisting of 139 research papers. 
First, we tokenized and lemmatized the titles and abstracts to standardize the text, by using NLTK\footnote{\url{https://www.nltk.org/}}. Next, we removed common English stop words (e.g., `the', `and') to eliminate noise. Finally, we filtered out query-specific and domain-related terms such as [`nlp', `natural', `language', `processing', `stereotype', `detection', 
`et al'], which were too frequent or generic to meaningfully contribute to the analysis. This allowed us to focus on the most distinctive terms in the corpus.

\begin{figure}[!ht]
\centering
\includegraphics[width=\textwidth]{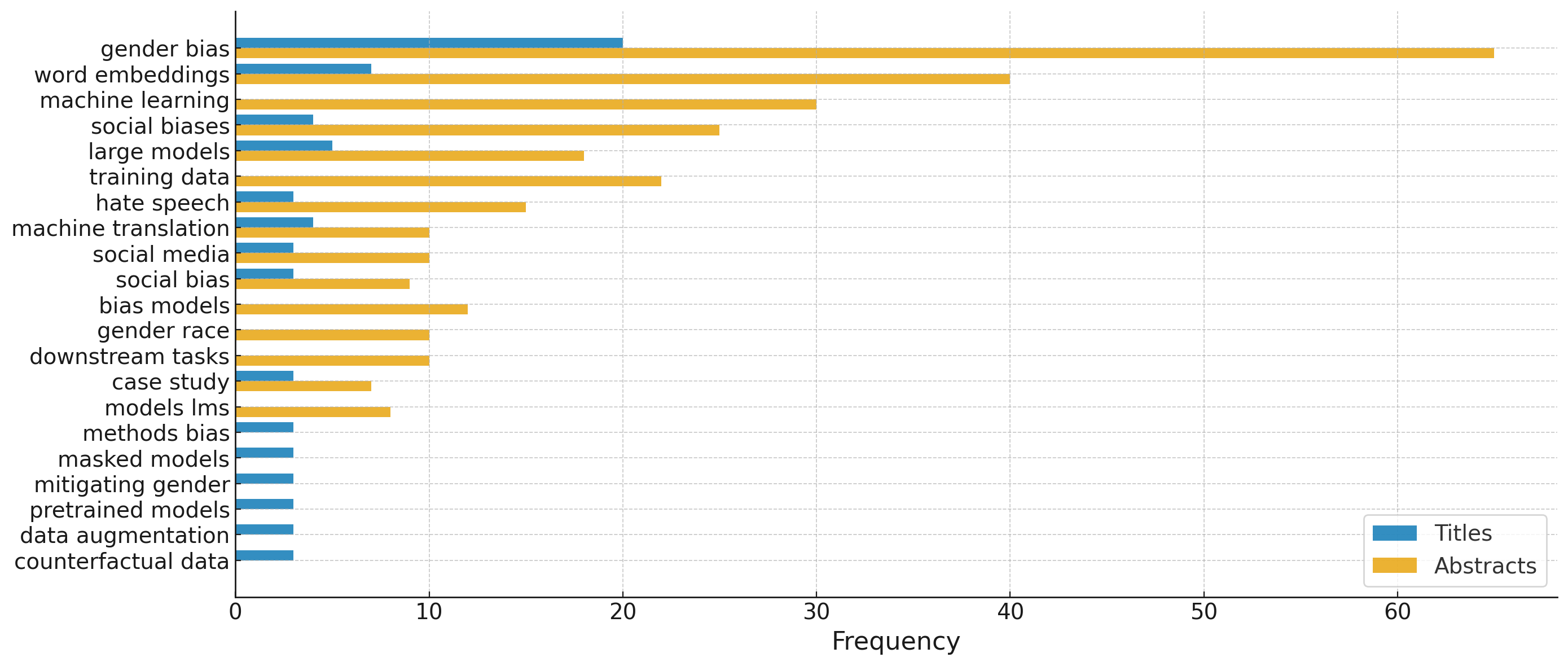}
\caption{Most frequent bigrams in Titles and Abstracts.\label{fig:bigrams-in-abstracts-titles}}\Description{}
\end{figure}

We were curious to analyze the textual content of Titles and Abstracts after removing the main query words to identify whether any interesting patterns would emerge regarding the relationship with \emph{Stereotype Detection}. For this purpose, we extracted the most frequent bigrams from both Titles and Abstracts. The results are presented in \autoref{fig:bigrams-in-abstracts-titles}.

\noindent The results highlight that \emph{gender bias} is overwhelmingly prevalent in our corpus of selected articles. This is hardly surprising, given the attention that gender bias has received in the field. Then we observe terms associated with the technical aspects of NLP, such as \emph{word embeddings}, \emph{large language models}, \emph{machine learning}, and \emph{training data}, as well as concepts related to societal phenomena like \emph{hate speech} or \emph{social biases}. The term \textit{machine translation} is also quite frequent, and, unsurprisingly, given the textual genre most NLP research focuses on, \emph{social media} also emerges as a frequently occurring bigram.

\begin{figure}[h]
\flushleft
\includegraphics[scale=0.39]{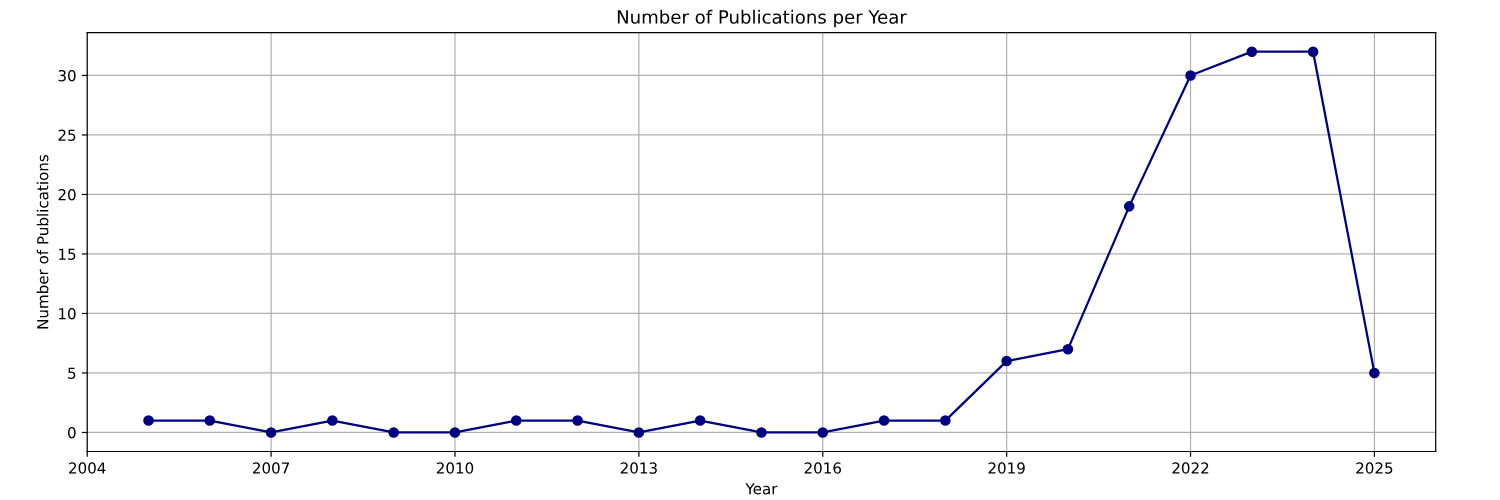}
\caption{Publication trend of \textit{`Stereotype Detection in NLP'} over time.\label{fig:publication-per-year}}\Description{}
\end{figure}

To examine the evolving interest in \textit{Stereotype Detection in NLP}, we further analyzed the occurrence of such terms over time, as illustrated in \autoref{fig:publication-per-year}. The data reveals a notable rise in the number of publications starting in 2019, with a significant surge in 2021. This upward trajectory continued sharply through 2023 and 2024, indicating a growing focus on the topic. Given this trend, we anticipate that research in this area will expand even further in the coming years.

Another indication of the field's novelty can be observed by examining its primary publishing venues. In \autoref{tab:venues}, we show the most prominent conferences and workshops where papers regarding \textit{Stereotype Detection} are currently being published.
The leading venue for papers of this topic is the open-access repository \textit{arXiv.org}, suggesting that much of the research in this area is either very recent or still in its pre-print stages. Unsurprisingly, other prominent venues include some of the main *CL conferences (ACL, EMNLP, NAACL and LREC). Consistent with the bigram analysis shown above, two other notable venues are the Conference on Fairness, Accountability, and Transparency (FAccT) and the Workshop on Gender Bias in Natural Language Processing (GeBNLP).

\noindent To conclude this section, we present a concise list of key papers, prioritizing the most influential ones -- specifically, those with over 50 citations (see \autoref{tab:most-cited-papers}). These highly cited works played a central role in our reading process, offering well-established insights, methodologies, and datasets in the field of \textit{Stereotype Detection}, which, despite its relatively recent history as an emerging  field in its own right.

\begin{table}[H]
\small
\begin{tabular}{lr}
\textbf{Venue} & \textbf{N\# papers} \\
\toprule
arXiv.org & 21 \\
Annual Meeting of the Association for Computational Linguistics (ACL) & 11 \\
Conference on Empirical Methods in Natural Language Processing (EMNLP)	& 8\\
North American Chapter of the Association for Computational Linguistics (NAACL) &	6\\
International Conference on Language Resources and Evaluation (LREC) &	5\\
Workshop on Gender Bias in Natural Language Processing (GEBNLP) &	4\\
Conference on Fairness, Accountability and Transparency (FAccT) &	3\\
AAAI/ACM Conference on AI, Ethics, and Society (AIES) &	3\\
Workshop on Online Abuse and Harms (WOAH) &	2\\
Conference on Lexical and Computational Semantics (*SEM) & 2\\
\bottomrule
\end{tabular}
\caption{Most frequent publishing venues.}\label{tab:venues}
\end{table}

\begin{table}[h]
\centering
\resizebox{\textwidth}{!}{
\begin{tabular}{llr}
\textbf{Reference} & \textbf{Title} & \textbf{Citations} \\
\toprule
\citet{nangia2020crows} & CrowS-Pairs: A Challenge Dataset for Measuring Social Biases in Masked Language Models & 587 \\ \midrule
\citet{sun2019mitigatinggenderbias} & Mitigating Gender Bias in Natural Language Processing: Literature Review & 513 \\ \midrule
\citet{kurita2019measuring} & Measuring Bias in Contextualized Word Representations & 421 \\ \midrule
\citet{parrish2022bbq} & BBQ: A hand-built bias benchmark for question answering & 292 \\ \midrule
\citet{deshpande2023toxicity} & Toxicity in ChatGPT: Analyzing Persona-assigned Language Models & 285 \\ \midrule
\citet{zmigrod2019counterfactual} & Counterfactual Data Augmentation for Mitigating Gender Stereotypes in Languages with Rich Morphology & 265 \\ \midrule
\citet{blodgett2021stereotyping} & Stereotyping Norwegian Salmon: An Inventory of Pitfalls in Fairness Benchmark Datasets & 261 \\ \midrule
\citet{liang2020towardsdebiasing} & Towards Debiasing Sentence Representations & 215 \\ \midrule
\citet{shaikh2023secondthought} & On Second Thought, Let’s Not Think Step by Step! Bias and Toxicity in Zero-Shot Reasoning & 152 \\ \midrule
\citet{rudinger2017social} & Social Bias in Elicited Natural Language Inferences & 132 \\ \midrule
\citet{nozza2021honest} & HONEST: Measuring Hurtful Sentence Completion in Language Models & 128 \\ \midrule
\citet{stanczak2021surveygenderbias} & A Survey on Gender Bias in Natural Language Processing & 98 \\ \midrule
\citet{papakyriakopoulos2020bias} & Bias in word embeddings & 95 \\ \midrule
\citet{chang2019biasandfairness} & Bias and Fairness in Natural Language Processing & 59 \\ \midrule
\citet{levy2021collecting} & Collecting a Large-Scale Gender Bias Dataset for Coreference Resolution and Machine Translation & 58 \\ \midrule
\citet{cryan2020detecting} & Detecting Gender Stereotypes: Lexicon vs. Supervised Learning Methods & 50 \\
\bottomrule
\end{tabular}
}
\caption{Most-cited papers with more than 50 citations.}\label{tab:most-cited-papers}
\end{table}

\subsection{Extra: Projects on GitHub and HuggingFace}
To gain a complete overview of the existing Stereotype Detection projects that are currently being developed in the NLP community, we searched GitHub and HuggingFace using the terms \textit{Stereotype Detection} in the platforms’ search engines. These queries yielded \textbf{24 hits on GitHub and 3 hits on HuggingFace} containing relevant content. The search for projects was last performed in February 2025. 

Most of the repositories found were created by participants of the shared task ``Profiling Irony and Stereotype Spreaders on Twitter (IROSTEREO)'' organized at PAN in 2022 \citep{ortega2021irostereo}, which focused on determining whether the author of a Twitter feed was keen on spreading irony and stereotypes towards immigrants, women and LGBT+ people. 
Similarly, other repositories belong to participants of the 2024 shared task ``DETESTS-Dis'' organized at IberLEF in 2024, to detect and classify explicit and implicit stereotypes in texts from social media and comments on news articles in Spanish, incorporating learning with disagreement techniques \citep{schmeisser2024overviewdetests}. 
The remaining repositories are dataset-related, which we will describe in greater detail in a later section (see Section \ref{subsec:dataset-corpora}).

\medskip

\noindent In the following section, starting from these foundational works, we transition into the main core of our literature review. By listing and analyzing these contributions, at first, we aim to establish a solid foundation for future research in \textit{Stereotype Detection}, highlighting existing advancements, identifying gaps and key challenges.
\section{Organized Overview of Literature on Stereotypes}\label{sec:main-literature-categories}
This section forms the core of the manuscript, presenting an extensive compilation of the research work we have reviewed. Our initial selection comprised the 50 most cited papers in the field, which we then expanded to include highly interconnected works frequently referenced within our starting database. The section is conceived as a \textit{vademecum} -- a reference guide offering a broad yet organized overview of research conducted thus far in Stereotype Detection. While not systematic, it provides a structured classification of the key research directions and subfields in NLP related to stereotypes. In \autoref{tab:categories-literature}, we propose a reasoned grouping of the findings from published studies.


\begin{table}[h]
\centering 
\footnotesize
\begin{tabular}{p{.02cm}p{2.5cm}p{5.2cm}p{5.8cm}}
& \textbf{Category} & \textbf{Description} & \textbf{References} \\
\toprule
1 & \textsc{Theoretical} & Papers focused on defining, conceptualizing, or analyzing theoretical frameworks related to stereotypes, bias, prejudice, or discrimination. These works typically provide a foundation for future empirical research or propose new perspectives.
Papers focused on operationalizing stereotypes. 
Including taxonomies and annotation schemes for stereotype detection. & \citet{cao2022theorygrounded,fiske1993controlling,fiske1998stereotyping,fiske2002modelmixed,fiske2024web,fraser2024howdoes,kirk2022handling,koch2016abc,koch2020groupswarmth,nicolas2022spontaneous,prabhumoye2021casestudy,sanchezjunquera2021yhowdoyou,schmeisser2022criteria} \\

\midrule

2 & \textsc{Dataset/Corpora} & Papers presenting new datasets, annotated corpora, or language resources, often designed to facilitate research on stereotypes, biases, or related topics. & \citet{bosco2023detecting,bourgeade2023multilingual,cignarella2024queereotypes,jha2023seegull,levy2021collecting,nangia2020crows,parrish2022bbq,neveol2022frenchcrows,sanguinetti2020haspeede,ortega2021irostereo,schmeisser2024overviewdetests,nozza2021honest,nadeem2021stereoset,sanchezjunquera2021yhowdoyou,schmeisser2024stereohoax} \\

\midrule

3 & \textsc{Data Analysis} & Papers applying methods to specific use cases or providing real-world analysis of stereotypes and biases. Papers analyzing specific datasets to uncover trends, patterns, or prevalence of stereotypes, biases, or abusive content. These studies often provide insights into real-world phenomena through data-driven methods. & \citet{deshpande2023toxicity,fraser2021understanding,locatelli2023crosslingual,mendelsohn2020dehumanization,rudinger2017social,fraser2024howdoes} \\ 

\midrule

4 & \textsc{Debiasing/Mitigation} & Papers proposing methods, techniques, algorithms, or frameworks to mitigate or reduce bias and stereotypes in data, language models, or outputs. These studies typically evaluate the effectiveness of such approaches. & \citet{birhane2023hate,blodgett2021stereotyping,blodgett2022responsible,cao2022theorygrounded,chang2019biasandfairness,davani2023socialstereotypes,fraser2022extracting,kurita2019measuring,liang2020towardsdebiasing,nadeem2021stereoset,nozza2021honest,pujari2022reinforcement,ranaldi2024trip,shaikh2023secondthought,shen2021contrastive,sun2019mitigatinggenderbias,ungless2022robust,vargas2023sociallyresponsible,zhou2023causaldebias,zmigrod2019counterfactual} \\

\midrule
5 & \textsc{Stereotype Detection \newline (Classification Task)} &  Papers focused on developing or utilizing computational methods for detecting stereotypes and/or biases. This category includes papers proposing or testing models and algorithms in classification tasks. & \citet{bosco2023detecting,cignarella2024queereotypes,pujari2022reinforcement,sanchezjunquera2021yhowdoyou,sanguinetti2020haspeede,schmeisser2024stereohoax,vargas2023sociallyresponsible} \\ 

\midrule

6 & \textsc{Other: Shared Tasks, \newline Projects, Tutorials, \newline Position Papers } & Papers that do not fit neatly into the other categories, such as descriptions of large-scale projects, overviews of shared tasks, or position papers discussing challenges in the field. & \citet{bhatt2022caseofindia,fortuna2018survey,sun2019mitigatinggenderbias,sanguinetti2020haspeede,blodgett2021stereotyping,stanczak2021surveygenderbias,derrico2024project,chang2019biasandfairness,ortega2021irostereo,schmeisser2024overviewdetests,schutz2021sexism} \\ 
\bottomrule
\end{tabular}
\caption{Summary of the surveyed papers, divided by main categories.}\label{tab:categories-literature}
\end{table}

\begin{figure}
\centering
\includegraphics[width=0.42\linewidth]{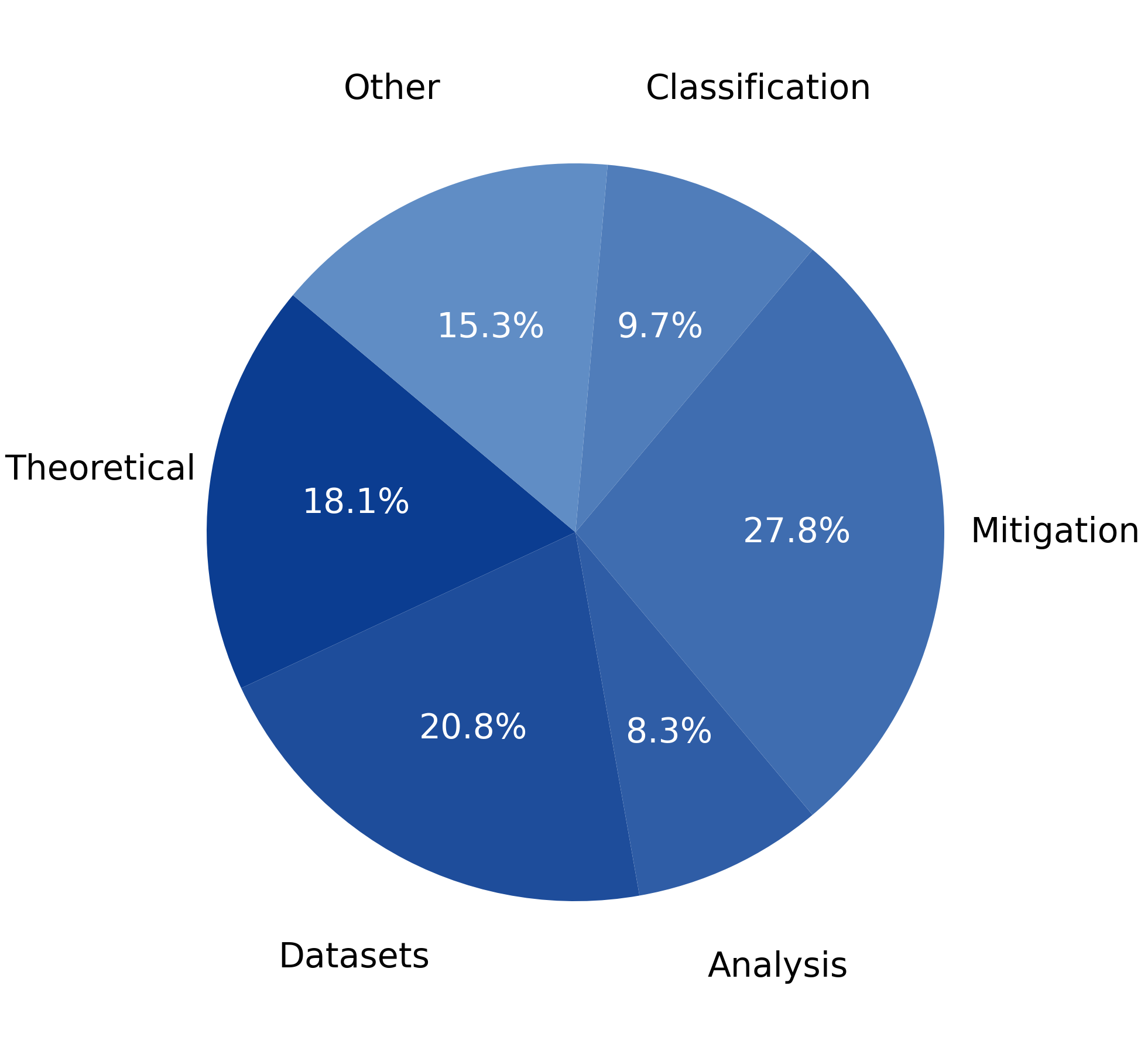}
\caption{Category distribution of the surveyed papers.}\label{fig:pie-chart-square}\Description{}
\end{figure}

\noindent The organization in \autoref{tab:categories-literature} includes \textit{theoretical contributions} defining and conceptualizing stereotypes, \textit{dataset creation} for stereotype research, \textit{data analysis} for real-world bias identification, \textit{debiasing and mitigation strategies}, \textit{stereotype detection} through computational methods, and \textit{miscellaneous contributions} such as shared tasks, tutorials, and position papers. Each category highlights key studies that contribute to understanding and addressing biases in NLP. {Furthermore,} \autoref{fig:pie-chart-square} {illustrates the distribution of papers within the six categories, highlighting the majority of papers belonging to Debiasing/Mitigation.}

The following subsections will delve deeper into each category, outlining their methodologies, findings, and implications for bias research. The primary objective of this section is to list prominent works rather than to analyze them; more in-depth commentary and discussion will follow in Section \ref{sec:challenges}.

\subsection{Theoretical Work\label{subsec:theoretical}}

Theoretical studies and frameworks are key in the field of stereotypes as they provide structured lenses to understand, identify, and analyze stereotypes within social, cultural, and linguistic contexts. Table~\ref{tab:theoretical} shows an overview {of the 13 papers} that we have identified and which focus on defining, conceptualizing, or analyzing theoretical frameworks related to stereotypes, bias, prejudice, or discrimination. 

\begin{table}[h]
\centering
\footnotesize
\begin{tabular}{p{2.8cm}p{8.1cm}p{3.1cm}}
\textbf{Reference} & \textbf{Overall Description} & \textbf{Phenomenon} \\
\toprule

\citet{fiske1993controlling} & Deals with how individuals in positions of power are more likely to rely on stereotypes when evaluating others, as power influences cognitive processing and reduces the complexity of social judgments. & (Gender) stereotypes. \\ 

\midrule
\citet{fiske1998stereotyping} & Presents how stereotypes are formed and how they shape biased attitudes and behaviors.  & Social stereotypes, prejudice, and discrimination across various social groups. \\
\midrule

\citet{fiske2002modelmixed} & Presents the Stereotype Content Model (SCM) which suggests that stereotypes are shaped by two primary dimensions: competence and warmth. & Stereotypes across different social groups. \\
\midrule

\citet{koch2016abc} & Extends stereotype content research by introducing the ABC model, which categorizes stereotypes along three dimensions: Agency (socioeconomic success), Beliefs (conservative-progressive ideology), and Communion (warmth and trustworthiness).  & Stereotypes across different social groups, such as race, gender, and political affiliation.\\
\midrule

\citet{koch2020groupswarmth} & Observes how perceptions of group warmth vary across individuals, highlighting that warmth judgments are shaped by personal experiences and social context rather than being universally consistent. & Various social groups (e.g., based on race, gender, or status).  \\
\midrule

\citet{prabhumoye2021casestudy} & Explores the application of deontological ethics — namely, the generalization principle and respect for autonomy via informed consent — within the field of NLP. & Various social groups. \\
\midrule

\citet{sanchezjunquera2021yhowdoyou} & Introduces a detailed taxonomy for classifying stereotypes about immigrants. It categorizes biases into key themes such as economic threats, cultural differences, criminalization, and victimization, helping to systematically analyze how immigrants are portrayed in social media discourse. & Stereotypes towards immigrants.\\
\midrule

\citet{cao2022theorygrounded} & Presents the sensitivity test (SeT) for measuring stereotypical associations from language
models.  & Stereotypes across various social groups. \\
\midrule

\citet{kirk2022handling} & Proposes an analytical framework categorizing harms on three axes: (1) the harm type; (2) whether a harm is sought as a feature of the research design; and (3) who it affects & Marginalized groups.\\
\midrule

\citet{nicolas2022spontaneous} & Proposes the spontaneous stereotype content model (SSCM) as an initial and comprehensive descriptive model with the aim of understanding the structure, properties, and predictive value of spontaneous stereotypes.& Stereotypes across various social groups.  \\
\midrule

\citet{schmeisser2022criteria} & Provides an operationalized definition for the annotation of new corpora by characterizing the different forms in which stereotypes appear. & Implicit stereotypes \newline (racial/ethnic). \\ 
\midrule 

\citet{fiske2024web} & This web contribution aims to deepen the understanding of biases in contemporary society by distinguishing between prejudice, stereotypes, and discrimination while exploring the differences between blatant and subtle forms of bias. & Prejudice, Stereotype, Discrimination, Bias.\\ 
\midrule

\citet{fraser2024howdoes} & Analyzes different data sources and different data collection methods regarding the study of stereotypes. & Stereotype (stereotypical beliefs common in the society). 
\\ 
\bottomrule
\end{tabular}
\caption{Theoretical work and papers presenting taxonomies or annotation schemes.\label{tab:theoretical}}
\end{table}

Early work on the field of stereotypes explored the cognitive processes behind prejudice and discrimination, and highlighted their impact on social interactions and group dynamics~\cite{fiske1998stereotyping}. One key area of interest was the reciprocal relationship between power and stereotyping~\cite{fiske1993controlling}. \citet{fiske1993controlling} argued that those in power tend to pay less attention to others, making them more prone to relying on stereotypes. Several papers also proposed theoretical models to understand stereotypes, their structure, and their social implications. \citet{fiske2002modelmixed} introduced the Stereotype Content Model (SCM), a model according to which stereotypes are systematically structured along two primary dimensions: warmth (how friendly or trustworthy a group is perceived as) and competence (how capable a group is perceived as). These perceptions are shaped by status and competition, meaning groups with high status are seen as competent, while those perceived as competitors are seen as cold. Expanding this model, \citet{koch2016abc} proposed adding a third dimension — agency (power or dominance) — arguing that stereotypes are also linked to socioeconomic success and political ideology. Similarly, \citet{koch2020groupswarmth} explored how warmth perceptions vary depending on individual experiences and social consensus, challenging the idea that warmth judgments are universal. Another theoretical contribution was made by \citet{nicolas2022spontaneous}, who explored how stereotypes form in different contexts, developing a taxonomy of stereotypes and demonstrating that they can emerge in various social and cultural settings, rather than being fixed.

Several studies expand beyond psychological models to examine stereotypes in the context of language and computational systems. For example, \citet{cao2022theorygrounded} and \citet{fraser2024howdoes} analyze how AI systems encode, reproduce, and even amplify social stereotypes. The intersection of stereotypes and NLP is also explored by \citet{kirk2022handling} and \citet{prabhumoye2021casestudy}, who emphasize the ethical implications of stereotype propagation in AI systems. \citet{prabhumoye2021casestudy} argue that bias in AI is not merely a technical issue but a moral responsibility, requiring NLP researchers to adopt ethical guidelines to minimize harm. Additionally, \citet{sanchezjunquera2021yhowdoyou} introduced the \textit{StereoImmigrants} dataset, along with a taxonomy for classifying stereotypes about immigrants, enabling a structured approach to identifying and analyzing biases in discourse.

\subsection{Dataset and Corpora}\label{subsec:dataset-corpora}

Table~\ref{tab:datasets} shows the papers that have proposed datasets containing stereotypes as one of the annotated dimensions. The developed datasets employ a variety of data types, including social media data, curated sentence pairs, synthetic data, and multilingual corpora, and aim to address different aspects of bias detection. All the datasets are ordered chronologically according to their release date, i.e. the publication that describes them.

\begin{table}[h]
\centering
\footnotesize
\begin{tabular}{p{3cm}p{1.9cm}p{3.2cm}p{2.6cm}p{2.3cm}}
\textbf{Reference} & \textbf{Language} & \textbf{Size/Source} & \textbf{Phenomena} & \textbf{Target/Characteristics} \\
\toprule

\citet{nangia2020crows} \newline \textsc{CrowS-pairs} & English & 1,508 minimally distant sentence pairs & Stereotypes & See Appendix \ref{appendixB} \\
\midrule

\citet{sanguinetti2020haspeede} \newline \textsc{HaSpeeDe2} & Italian & 8,602 tweets and news & Hate speech, stereotype and nominal utterances & Muslims, Roma and immigrants  \\
\midrule

\citet{levy2021collecting} \newline \textsc{BUG} & English & 108K sentences sampled from Wikipedia, PubMed abstracts, and Covid19 research papers & Gender bias & Gender  \\ 
\midrule

\citet{nadeem2021stereoset} \newline \textsc{StereoSet} & English & Sentences triplets & 1) stereotypical, \newline 2) anti-stereotypical, \newline 3) unrelated association & Gender, profession, \newline race, and religion  \\
\midrule 

\citet{nozza2021honest} \newline \textsc{Honest} & English, Italian, French, Portuguese, Romanian, Spanish  & 420 manually-created sentences for each language & Stereotypes & Gender  \\ 
\midrule

\citet{ortega2021irostereo} \newline \textsc{IROSTEREO} & English & posts from 600 Twitter users & Irony and stereotype spreaders & See Appendix \ref{appendixB} \\ 
\midrule

\citet{sanchezjunquera2021yhowdoyou} \newline \textsc{StereoImmigrants} &  Spanish & 3,635 sentences extracted from transcriptions of the Spanish parliament speeches & Stereotypes & Immigrants \\
\midrule

\citet{neveol2022frenchcrows} \newline \textsc{French CrowS-pairs} & French & 1,677 sentence pairs (1467 translated from English Crows-Pairs + 210 new in French + backtranslated to English) & Stereotypes &  See Appendix \ref{appendixB} \\ 
\midrule 

\citet{parrish2022bbq} \newline \textsc{BBQ} & English & 58,492 unique examples of sentences and paragraphs & Biases & See Appendix \ref{appendixB}  \\
\midrule

\citet{bosco2023detecting} \newline \textsc{FB-Stereotypes} & Italian & 2,990 Facebook posts & Proself/prosocial, hate speech, stereotype, prejudice, discredit & Immigrants  \\ 
\midrule 

\citet{bourgeade2023multilingual} \newline \textsc{Multilingual Racial \newline Hoaxes Corpus (MRHC)} & Italian, Spanish, French & 16,906 tweets & Racial stereotypes, Contextuality, Implicitness, Forms of discredit 
& Race  \\ 
\midrule

\citet{jha2023seegull} \newline \textsc{SeeGULL} & English & 7,750 (identity+attribute) pairs & Stereotypes & Geographical identity  \\ 
\midrule 

\citet{cignarella2024queereotypes} \newline \textsc{Queereotypes} & Italian & 3,427 tweets and 2,888 Facebook status + comment pairs & Hate Speech, aggressiveness, offensiveness, irony, stereotype and stance & LGBTQIA+  \\
\midrule 

\citet{schmeisser2024stereohoax} \newline \textsc{StereoHoax} & Italian, Spanish, French & 17,814 tweets & Stereotype, implicitness, contextuality, discredit & Immigrants \\
\midrule

\citet{schmeisser2024overviewdetests} \newline \textsc{DETESTS-Dis} & Spanish & 12,111 comment sentences and tweets & Stereotype, implicitness & Race  \\ 
\bottomrule

\end{tabular}
\caption{Papers presenting datasets, corpora and resources}\label{tab:datasets}
\end{table}

A predominant feature among the datasets presented in the table is the presence of English-centric datasets, such as \textsc{CrowS-Pairs} \cite{nangia2020crows}, \textsc{StereoSet} \cite{nadeem2021stereoset}, \textsc{BBQ} \cite{parrish2022bbq}, \textsc{SeeGULL} \cite{jha2023seegull}, \textsc{BUG} \cite{levy2021collecting}, and \textsc{IROSTEREO} \cite{ortega2021irostereo}. {In particular, 6 out of 15 datasets focus only on English language}
. While the focus on English is expected due to the availability of large-scale corpora, efforts to expand bias detection to other languages are evident. Datasets such as \textsc{Honest} \cite{nozza2021honest} span multiple Romance languages, while specific resources like \textsc{FB-Stereotypes} \cite{bosco2023detecting}, \textsc{MRHC} \cite{bourgeade2023multilingual}, \textsc{Queereotypes} \cite{cignarella2024queereotypes}, and \textsc{StereoImmigrants} \cite{sanchezjunquera2021yhowdoyou} introduce bias-related corpora in Italian, Spanish, and French. {However, a noticeable gap persists in the availability of comparable datasets for many other languages, particularly low-resource and non-European ones.}

The datasets listed vary significantly in size, ranging from a few hundred manually created instances (e.g., \textsc{honest} with 420 instances per language) to large-scale corpora containing tens of thousands of sentences (e.g., \textsc{bbq} with 58,492 examples and \textsc{bug} with 108K sentences). {From the table we notice that two datasets contain less than 1,000 sentences.} Medium-sized datasets, such as \textsc{Queereotypes} (6,315 texts), \textsc{FB-stereotypes} (2,990 Facebook posts), and \textsc{StereoImmigrants} (3,635 manually annotated sentences), provide focused, yet substantial corpora for analysis. The multilingual datasets, such as \textsc{mrhc} and \textsc{StereoHoax}, contain fewer more than 15,000 instances, ensuring coverage across different languages.

In terms of the linguistic phenomena captured, {13 out of 15 datasets} focus on stereotypes, as seen in \textsc{CrowS-Pairs}, \textsc{SeeGULL}, \textsc{Honest}, and \textsc{StereoHoax} \cite{schmeisser2024stereohoax}, with others addressing broader biases, such as BBQ. Additional considerations include irony and stereotype spread in \textsc{IROSTEREO} and proself versus prosocial discourse in \textsc{FB-Stereotypes}. {In addition, there are 3} datasets, \textsc{MRHC}, \textsc{DETESTS-Dis} and \textsc{StereoHoax}, which emphasize implicitness and contextuality, underscoring the challenge of detecting biases that are not overtly expressed. The intersection of bias detection with hate speech, offensiveness, and stance analysis is also evident in {4 datasets} like \textsc{Queereotypes}, \textsc{HaSpeeDe2}, and \textsc{StereoImmigrants}.

The datasets focus on various biases, stereotypes, and discriminatory content across multiple domains, including gender, race, profession, religion, immigration, and LGBTQIA+ identities. Some datasets, such as \textsc{CrowS-pairs} and \textsc{StereoSet}, specifically target stereotypical associations, while others, like \textsc{HaSpeeDe2}, \textsc{Queereotypes}, and \textsc{StereoImmigrants}, also analyze hate speech, offensiveness, and stance. The StereoImmigrants dataset is particularly notable for its structured taxonomy of stereotypes about immigrants, covering themes such as criminality, economic impact, and cultural integration. Additionally, datasets like \textsc{IROSTEREO} and \textsc{MRHC} examine the role of irony, implicitness, and contextuality in spreading stereotypes. {We found 3 out of 15 datasets} that were multilingual, including \textsc{Honest}, \textsc{StereoHoax}, and \textsc{StereoImmigrants}, which enable cross-linguistic comparisons of bias. Further details on the specific characteristics of each dataset can be found in \autoref{appendixB}.

In this section, we provided a qualitative and quantitative overview of the datasets, while in Section \ref{sec:challenges}, we adopt a more critical perspective to explore their characteristics and areas for further development.

\subsection{Computational Analyses of Stereotypes\label{subsec:computational-analyses}}

{The 6 studies listed} in \autoref{tab:computational-analyses} all present some kind of analyses of stereotypes by using computational techniques.

\begin{table}[h]
\centering
\footnotesize
\begin{tabular}{p{2.6cm}p{2.7cm}p{4.5cm}p{3.6cm}}
\textbf{Reference} & \textbf{Language} & \textbf{Goal} & \textbf{Phenomenon} \\
\toprule

\citet{rudinger2017social} & English 
& The main goal of the paper is to investigate and quantify the social biases that emerge in natural language inference data as a result of the human elicitation process. & Social bias across various targets. \\
\midrule

\citet{mendelsohn2020dehumanization} & English & The main goal of this paper is to propose a systematic, computational framework that allows researchers to analyze dehumanizing language. & Dehumanization (with a specific focus on LGBTQIA+ people). \\
\midrule

\citet{fraser2021understanding} & English \newline (U.S. context) & The main goal of this paper is to leverage the Stereotype Content Model (SCM) to develop computational methods that both detect and mitigate stereotypical language. & Stereotypes (SCM) across 79 different target groups. \\
\midrule

\citet{locatelli2023crosslingual} & German, French, English, Italian, Spanish, Portuguese and Norwegian & The paper’s main goal is to investigate how homotransphobia manifests across different cultural and linguistic contexts on Twitter. & LGBTQIA+, Cultural stereotypes, and Homotransphobia.\\
\midrule

\citet{deshpande2023toxicity} & English & The paper aims to investigate how assigning different personas to a language model (in this case, ChatGPT) influences the level and nature of toxic language it generates. & Toxicity towards various job titles regarding gender, race, sexual orientation, country, profession, religion, name, political organization, government type.\\
\midrule

\citet{fraser2024howdoes} & English \newline (U.S. context) & The goal of the paper is to investigate how the content of stereotypes varies depending on the data source used to derive them. & Politicians, teachers, CEOs, scientists, bankers, accountants, engineers, farmers, lawyers, and nurses. \\

\bottomrule

\end{tabular}
\caption{Papers presenting computational analyses of stereotypes.}\label{tab:computational-analyses}
\end{table}

For instance, \citet{rudinger2017social} examine biases introduced through human data elicitation in Natural Language Inference tasks, specifically analyzing the SNLI dataset \citep{bowman2015large}. The authors demonstrate how hypotheses generated by crowd workers can amplify stereotypical associations (e.g., gender, race, or age biases). They employ both statistical measures, such as pointwise mutual information (PMI), and qualitative analyses to illustrate how biases emerge and influence NLP models. \citet{fraser2021understanding}, in contrast, apply the Stereotype Content Model to computationally detect and mitigate stereotypes across 79 target groups. Their approach seeks to quantify these associations and develop methods for countering their negative impact on NLP applications.

Other studies extend beyond the computational analysis of stereotypes into specific manifestations of harmful language. \citet{mendelsohn2020dehumanization} integrate social psychology theories with computational linguistic methods to analyze dehumanizing language (particularly targeting the LGBTQIA+ community). The authors propose a systematic framework that operationalizes dehumanization, identifying, quantifying, and interpreting linguistic markers of dehumanizing discourse. Similarly, \citet{locatelli2023crosslingual} investigate homotransphobia across multiple languages and cultural contexts, analyzing queer-related discourse on Twitter/X in seven different languages. In addition to assessing the prevalence of homotransphobic content, the authors introduce a taxonomy to classify public discourse surrounding LGBTQ+ issues, providing insights to inform hate speech detection strategies. On the theoretical aspects, see also Section \ref{subsec:theoretical}.

A recent paper from \citet{fraser2024howdoes} investigated how the content of stereotypes varies depending on the data source used to derive them. In essence, the paper compares stereotype content extracted from different corpora (spontaneous adjectives \cite{nicolas2022spontaneous}, crowd-sourced StereoSet \citep{nadeem2021stereoset}, ChatGPT, Twitter/X), analyzing whether and how patterns of associations shift across sources. This work highlights the critical role of data selection in shaping our understanding of societal stereotypes, emphasizing the need for context-aware models that account for variations in stereotype expression across different data sources.

Another work that includes ChatGPT for studying bias in AI-generated text is that from \citet{deshpande2023toxicity}, in which the authors examine how attributing specific personality traits to ChatGPT can influence the level of toxicity in its responses. Their study quantifies how different persona assignments vary the model’s language outputs, providing insights that could inform mitigation strategies for harmful language in AI-driven systems.

\subsection{Stereotype Mitigation and Debiasing\label{subsec:debiasing-mitigation}}

The papers on the mitigation of stereotypes and biases constitute the broader category in our literature review {(20 papers)}. In \autoref{tab:debiasing-mitigation-0}, we enumerate all those that belong to this category along with the languages included in the study and the framing of the specific phenomenon. Papers are listed chronologically and then alphabetically.

\begin{table}[h]
\centering
\footnotesize
\begin{tabular}{p{2.5cm}p{3.5cm}p{8cm}}
\textbf{Reference} & \textbf{Language} & \textbf{Phenomenon} \\
\toprule
\citet{chang2019biasandfairness} & English & {Gender bias, racial bias, and social biases in NLP models.} \\ \midrule
\citet{kurita2019measuring} & English & Bias in embeddings (mostly occupational). \\ \midrule
\citet{sun2019mitigatinggenderbias} & English & Gender bias in NLP models and word embeddings. \\ \midrule
\citet{zmigrod2019counterfactual} & Spanish, Hebrew, French, Italian & Gender bias / stereotyping in languages with rich morphology. \\ \midrule
\citet{liang2020towardsdebiasing} & English & Sentence-level social biases (gender, race, and religion). \\ \midrule
\citet{blodgett2021stereotyping} & English & Stereotyping and social biases in NLP benchmark datasets. \\ \midrule
\citet{nadeem2021stereoset} & English & Stereotypical biases (gender, profession, race, religion). \\ \midrule
\citet{nozza2021honest} & English, Italian, Portuguese, \newline French, Romanian, Spanish & Hurtful stereotypes and biases related to gender roles. \\ \midrule
\citet{shen2021contrastive} & English & Biases in tasks such as sentiment analysis, hate speech detection, \newline profession classification, activity recognition.\\ \midrule
\citet{blodgett2022responsible} & Language agnostic & Fairness, transparency, justice, and ethics of computational systems. \\ \midrule
\citet{cao2022theorygrounded} & English (U.S.) & Stereotypical associations related to social groups and intersectional identities. \\ \midrule
\citet{fraser2022extracting} & English & Age-related stereotypes and over-generalizations. \\ \midrule
\citet{pujari2022reinforcement} & English & Stereotype (explicit, implicit and non-stereotype). \\ \midrule
\citet{ungless2022robust} & English & Stereotypical biases, ``cold'' and ``incompetent'' traits. \\ \midrule
\citet{birhane2023hate} & English & Hateful content and societal biases. \\ \midrule
\citet{davani2023socialstereotypes} & English & Social stereotypes and biases in hate speech annotation and classification \newline processes. \\ \midrule
\citet{shaikh2023secondthought} & English & Social biases and toxic content. \\ \midrule
\citet{vargas2023sociallyresponsible} & Portuguese, English & Social stereotypes and biases in hate speech detection systems. \\ \midrule
\citet{zhou2023causaldebias} & English & Demographic biases and social stereotypes. \\ \midrule
\citet{ranaldi2024trip} & English & Biases related to gender, race, religion, and profession.\\
\bottomrule
\end{tabular}
\caption{Papers presenting approaches and techniques for stereotype mitigation and for debiasing language models.\label{tab:debiasing-mitigation-0}}
\end{table}

Among the work reviewed, \citet{chang2019biasandfairness} provide a {{comprehensive survey}} of bias and fairness in NLP, examining historical contexts, bias quantification methodologies, and mitigation strategies. Their study spans various NLP applications, including word embeddings, coreference resolution, machine translation, and vision-and-language tasks. Similarly, \citet{sun2019mitigatinggenderbias} analyze gender bias in NLP, categorizing mitigation strategies into pre-processing, in-processing, and post-processing methods while also highlighting challenges.

As seen in Section \ref{subsec:dataset-corpora}, {{datasets}} play a crucial role in measuring and evaluating biases in NLP models. For instance, \citet{nadeem2021stereoset} introduce StereoSet, a dataset specifically designed to assess stereotypical biases in pretrained language models across gender, profession, race, and religion using the Context Association Test. These evaluations underscore the limitations of current datasets in capturing and quantifying bias effectively, necessitating more refined methodologies for stereotype assessment.
However, dataset creation can often pose some criticalities. Starting from this point of view, \citet{blodgett2021stereotyping} audit four widely used benchmark datasets ({StereoSet}, CrowS-Pairs WinoBias, and Winogender), identifying key pitfalls such as ambiguous definitions and assumptions that undermine their effectiveness in quantifying stereotyping. 

In a similar direction, several other studies have focused on the {{quantification of bias in NLP models}}. Among these, the work of \citet{kurita2019measuring} proposes a template-based approach to assess bias in BERT, particularly in occupational contexts, demonstrating how biases embedded in pre-trained language models could influence downstream tasks. Secondly, \citet{liang2020towardsdebiasing} introduce a post-hoc method aimed at reducing social biases in sentence representations, targeting gender, race, and religious biases, while \citet{cao2022theorygrounded} 
analyze group-trait associations in language models, employing the \textit{sensitivity test} to compare model outputs against human judgments. Finally, \citet{nozza2021honest} propose \textit{HONEST}, a systematic evaluation framework that utilizes lexicon-based methodologies for recognizing stereotypes in language model completions across multiple languages, with a particular emphasis on gender-specific biases.

Among the many relevant research papers, we have selected works that present actual {{computational methodologies}} for bias mitigation in NLP models.
For instance, \citet{zmigrod2019counterfactual} introduce a counterfactual data augmentation (CDA) approach to mitigate gender biases in morphologically rich languages by converting masculine-inflected sentences into their feminine counterparts and vice versa.
\citet{shen2021contrastive} propose a contrastive learning framework designed to promote similarity among instances sharing the same class label while increasing separation between those differing in protected attributes; and  \citet{ungless2022robust} build on this by applying the SCM to modify contextualized word embeddings, ensuring stereotype reduction without compromising overall model performance.
Some work focuses on {{debiasing at the fine-tuning level or at the training level}}, such as \citet{zhou2023causaldebias}, who propose \textit{Causal-Debias}, a framework that integrates debiasing directly into the fine-tuning process of pretrained language models by isolating non-causal bias-related factors while preserving task-relevant information. \citet{ranaldi2024trip} investigate biases in large language models, specifically LLaMA and OPT, and apply Low-Rank Adaptation (LoRA) to effectively reduce biases related to gender, race, religion, and profession. \citet{pujari2022reinforcement} introduce a reinforcement learning-based approach for detecting and mitigating both explicit and implicit stereotypes in language models, improving model fairness during training and inference.

Other relevant work that we have included in this section, focuses on bias and its intertwined relationship with {{implicit hate speech}}. Among others, \citet{davani2023socialstereotypes} highlight how annotators' biases influence hate speech detection, revealing that classifiers often reinforce societal stereotypes against marginalized groups. Secondly, \citet{vargas2023sociallyresponsible} propose an analysis method to assess bias in hate speech classification systems, particularly in English and Portuguese datasets from social media platforms. Additionally, \citet{shaikh2023secondthought} explore the effects of zero-shot Chain of Thought (CoT) prompting in large language models, demonstrating that CoT often exacerbates bias and toxicity in model outputs.

To conclude, beyond all these technical contributions, other researchers have explored the more ethical dimensions of this field of research. \citet{blodgett2022responsible} advocate for a human-centered approach to NLP fairness, drawing on insights from Human-Computer Interaction to mitigate potential harms in language technologies.

\subsection{Stereotype Detection as a Classification Task\label{subsec:stereotype-detection}}

\autoref{tab:stereotype-detection-classification} presents an overview of various studies on stereotype detection across different languages, models, and methodologies. Each row represents a research paper, detailing the language(s) analyzed, the models employed, and the corresponding metrics used to assess the performance.

\begin{table}[h]
\centering
\footnotesize
\begin{tabular}{p{3cm}p{2.6cm}p{4cm}p{4cm}}
\textbf{Reference} & \textbf{Language} & \textbf{Models / Methodology} & \textbf{Performance Metrics} \\
\toprule

\citet{sanguinetti2020haspeede} & Italian & BERT, AlBERTo and UmBERTo, \newline XLMRoBERTa, DBMDZ, Bi-
LSTM, \newline Rule-based classifiers. & \textsc{{Best scores in the shared task}}: \newline Macro F1: 0.770 (Twitter) \newline Macro F1: 0.720 (News data) \\
\midrule

\citet{sanchezjunquera2021yhowdoyou} & Spanish & M-BERT, XLM-RoBERTa, BETO, SpanBERTa leveraging  Pointwise Mutual Information. & M-BERT, Accuracy: 0.829 \newline 
XLM-RoBERTa, Accuracy: 0.780 \newline
BETO, Accuracy: 0.861 \newline
SpanBERTa, Accuracy: 0.766  \\
\midrule

\citet{pujari2022reinforcement} & English &  Multi-task learning (MTL) model and Reinforcement Learning guided multi-task learning model (RL-MTL). & \textsc{{coarse-grained and fine-grained}}: \newline Baseline, F1: 0.657 and 0.614 \newline MTL, F1: 0.683 and 0.650 \newline RL-MTL, F1: 0.742 and 0.679 \\
\midrule

\citet{bosco2023detecting} & Italian & 10-fold cross-validation with SVM-RBF for baselines. GilBERTo with the addition of lexically relevant information. & Baseline SVM-RBF, F1: 0594 \newline
GilBERTo base, F1: 0.631 \newline
GilBERTo + lexical info, F1: 0.661 \\
\midrule

\citet{vargas2023sociallyresponsible} & Brazilian Portuguese, \newline English & BERT and fastText combined with different feature sets based on: Social Stereotype Analysis, bag-of-words,  offensive lexicon. & \textsc{{Reporting only the best scores \newline for each language setting}:} \newline Portuguese, F1: 0.860, \newline English, F1: 0.780 \\
\midrule

\citet{cignarella2024queereotypes} & Italian & mBERT, AlBERTo in base setting and expanded setting with extra data from the  Italian Hate Speech detection task. & \textsc{{Reporting only the best scores}:} \newline mBERT, F1: 0.735, \newline AlBERTo, F1: 0.744 \\
\midrule

\citet{schmeisser2024stereohoax} & Italian, Spanish, French & GilBERTo, BETO and CamemBERT base + second setting enhanced with contextual information. & \textsc{{Reporting only the best scores}:} 
\newline Italian, Macro F1: 0.805 \newline 
Spanish, Macro F1: 0.826 \newline
French, Macro F1: 0.716 \\
\bottomrule

\end{tabular}
\caption{Summary of classification experiments in selected studies.\label{tab:stereotype-detection-classification}}
\end{table}

Several studies focus on Italian, such as \citet{sanguinetti2020haspeede}, which presents the results of the shared task on Hate Speech Detection (HaSpeeDe2), where subtask B is Stereotype Detection. The teams participating in the competition employ a diverse set of classifiers including BERT variants (AlBERTo, UmBERTo, XLMRoBERTa), and the best scores achieved are: macro F1 of 0.770 (for the Twitter dataset) and 0.720 (for News dataset). Also in Italian, \citet{bosco2023detecting} explore an SVM-RBF baseline and the GilBERTo model, incorporating lexical features to improve performance, reaching an F1 score of 0.661. \citet{cignarella2024queereotypes} further investigate Italian stereotype detection with mBERT and AlBERTo, reporting F1 scores of 0.735 and 0.744, respectively. Meanwhile, \citet{schmeisser2024stereohoax} extend the analysis to Italian, Spanish, and French, using GilBERTo, BETO, and CamemBERT, with the best macro F1 scores reaching 0.805 (Italian), 0.826 (Spanish), and 0.716 (French).

For Spanish, \citet{sanchezjunquera2021yhowdoyou} explore several transformer models, including M-BERT, XLM-RoBERTa, BETO, and SpanBERTa, achieving the highest accuracy of 0.861 with BETO. The inclusion of BETO, a Spanish-specific model, demonstrates the importance of leveraging language-specific embeddings for improved performance.

In English, \citet{pujari2022reinforcement} employs multi-task learning (MTL) and reinforcement learning-enhanced MTL (RL-MTL), showing a progressive improvement in F1 scores from a baseline (0.657 coarse-grained, 0.614 fine-grained settings) to RL-MTL (0.742 and 0.679, respectively).

\citet{vargas2023sociallyresponsible} explore Brazilian Portuguese and English, integrating BERT and fastText models with feature sets based on a new methodology named Social Stereotype Analysis, bag-of-words, and the addition of offensive lexicons. The best F1 scores are 0.860 for Portuguese and 0.780 for English.

\subsection{Miscellaneous work on Stereotype Detection\label{subsec:others}}

Beyond the primary categories of theoretical works, datasets and corpora, analyses, methodologies for mitigation and debiasing, and classification framework, several works contribute to the broader discourse on stereotype detection by presenting large-scale projects, shared task overviews, tutorials, and position papers. These studies often serve as foundational discussions on challenges in bias detection, ethical concerns, and emerging trends in the field. For instance, \citet{bhatt2022caseofindia} and \citet{fortuna2018survey} provide comprehensive overviews of fairness and hate speech detection in NLP, while \citet{sun2019mitigatinggenderbias} and \citet{stanczak2021surveygenderbias} focus specifically on gender bias and strategies for its mitigation.

Shared task descriptions, such as \citet{sanguinetti2020haspeede} and \citet{schmeisser2024overviewdetests}, outline benchmark competitions designed to advance hate speech and stereotype identification across multiple domains. Similarly, projects like \citet{derrico2024project} and \citet{ortega2021irostereo} highlight interdisciplinary approaches to stereotype detection, incorporating insights from linguistics, psychology, and computational ethics. Position papers such as \citet{blodgett2021stereotyping} and \citet{chang2019biasandfairness} critically examine the limitations of existing work on stereotypes/biases, taking into account datasets, models, and emphasizing the need for improved annotation frameworks and fairer evaluation metrics.

\medskip

\noindent In this section, we provided a qualitative overview of the papers reviewed for our literature analysis. To facilitate clarity and accessibility, we have structured key information into tables, enabling readers to efficiently reference aspects such as the most prevalent theoretical frameworks, the scope of prior research, the languages of the studies, the models employed, and the corresponding results.  
In the following section (\S \ref{sec:challenges}), we examine the challenges that emerged from our analysis. Our findings suggest that this field presents significant opportunities for further development, and we will outline key areas where advancements are due.
\section{CHALLENGES\label{sec:challenges}}
Despite significant advancements in the field of \textit{Stereotype Detection} within the domain of Natural Language Processing, several challenges still exist that hinder the development of fairer and unbiased models. In this section, we explore the key points we discovered during our literature review and briefly discuss them.

\subsection{Consensus on Definitions and Theoretical Frameworks}
Throughout our survey of over 50+ papers on stereotype detection in NLP, we observed a general consensus on the definition of stereotypes. Most studies align with the foundational work of social psychologists such as \citet{fiske2002modelmixed} and \citet{koch2016abc}, whose frameworks dominate the conceptual landscape. However, other contributions, such as the work conducted by \citet{abele2007agency}, remain largely absent from being referenced to.  

Despite this apparent agreement on what constitutes a stereotype, we found a notable lack of clarity when it comes to distinguishing stereotypes from related concepts such as \textit{biases} and \textit{prejudice}. Most of the studies we reviewed tend to use these terms interchangeably, with no explicit attempt to delineate their boundaries. This fuzziness is likely due to the inherently subjective nature of stereotypes, which depend on socio-cultural context and evolve over time. As a result, while theoretical definitions exist, operationalizing them in computational terms remains a significant challenge.  

Also, we observed that theoretical studies on stereotypes often avoid focusing on specific target groups but aim for a broader range of attributes. This contrasts with the rest of the papers that we reviewed, which often frame their contributions as addressing stereotypes in a general sense, but later focus on highly specific instances, often restricted to particular datasets, a specific language or a very limited number of targets.

\subsection{Attribute Selection}\label{sec:attribute-selection}
Throughout our survey, we noticed that most computational methods rely on predefined attributes to operationalize stereotype detection. This targeted approach allows for a more precise identification of stereotypes but comes at the cost of reduced generalizability, as models tend to be optimized for specific biases rather than addressing stereotypes in a broader, more inclusive manner.  

Our literature review reveals that a significant portion of NLP research on stereotype detection centers around gender and racial stereotypes, with gender often analyzed in conjunction with occupational biases. However, as also noted by \citet{stanczak2021surveygenderbias} and by \citet{sun2019mitigatinggenderbias} in their surveys, most studies still adhere to a binary representation of gender (male/female), failing to represent non-binary identities, with the exclusion of the notable work of \citet{cao2022theorygrounded,deshpande2023toxicity} and \citet{jha2023seegull}.

Other important social attributes, such as age and sexual orientation, receive little, if any, attention in the literature with few exceptions \cite{fraser2022extracting}. Moreover, the concept of \textbf{\textit{\ul{intersectionality}}} \cite{crenshaw1989intersectionality}, which refers to the interconnected nature of social categorizations such as race, gender, and class in creating overlapping systems of discrimination or disadvantage, remains, even to date, largely unexplored with only a few notable examples \cite{tan2019assessing,subramanian2021evaluating,lalor2022benchmarking}.

\subsection{Datasets}\label{challenge:datasets}

Data collection and data quality are key factors for developing robust computational methods for stereotype detection. In total, {we identified 15 papers} that have developed a dataset for the task of stereotype detection. Some of those studies focus only on stereotypes~\cite{nangia2020crows,nadeem2021stereoset} while others address this problem as part of a broader task, such as {hate speech detection}~\cite{sanguinetti2020haspeede,cignarella2024queereotypes}. 
The datasets may vary in terms of the source/type of textual data, size, language, phenomenon, etc.
\medskip

\noindent \textit{\textbf{$\bullet$ Source / Type.}} The choice of data types is a very important factor in effectively measuring and mitigating stereotypes in NLP models. The datasets that we have found contain text from social media, curated text and synthetic data. In particular, {8 out of 15 proposed datasets} rely on text from social media (e.g., tweets, Facebook posts or Reddit). As this type of data is taken from the real-world, it is often unstructured and context-dependent, containing noisy and informal language. 

Another limitation of using content from social media is that data may not fully represent the diversity of stereotypes present in broader societal discourse, as platforms have different user demographics and content moderation policies. It could be interesting to incorporate a wider range of digital platforms, such as YouTube or TikTok, to capture a more comprehensive spectrum of biased discourse.

Other researchers use sentence pairs to measure the stereotypes in language models~\cite{nangia2020crows,nadeem2021stereoset}.~\citet{jha2023seegull} leverages generative models to create synthetic examples of stereotypes. Although this is an efficient way to create datasets, synthetic data can introduce challenges in ensuring data quality.

~\citet{CedricDeslandes2024} {discuss two risks of using synthetic data. The first risk is the illusory assurance associated with synthetic data, which refers to creating the illusion of representational fairness while neglecting deeper, cultural or qualitative disparities. The second risk is the consent circumvention that refers to violating ethical and legal frameworks built around data consent.}

\medskip

\noindent \textit{\textbf{$\bullet$ Phenomenon / Target.}} The analysis of phenomena modeled reveals a strong emphasis on hate speech, which appears most frequently among the surveyed papers. This suggests that a significant portion of research in this domain focuses on detecting and mitigating harmful language directed at individuals or groups. The presence of stereotypes, bias, and prejudice as recurring topics further highlights the growing interest in understanding how language and discourse contribute to societal inequalities and discrimination. The intersection of these phenomena suggests that many studies are not solely focused on a single aspect of offensive speech; rather, they explore multiple overlapping issues, such as how dataset bias can reinforce stereotypical narratives.

Another noteworthy pattern is the variety in categorization approaches used by different papers. While some explicitly label their research under broad terms like \textit{hate speech} or \textit{stereotype}, others employ more nuanced descriptions, such as \textit{homotransphobia}, \textit{aggressiveness}, and \textit{implicit bias}. This diversity in terminology underscores the complexity of the field and the challenge of establishing universally accepted definitions. The presence of multi-label classifications in several entries suggests an awareness of the interconnected nature of these linguistic phenomena, reinforcing the idea that addressing only one aspect might not be sufficiently accurate (see also Section \ref{sec:attribute-selection}).

A closer examination of target groups reveals a strong focus on gender biases, as evidenced in \textsc{BUG} \cite{levy2021collecting}, \textsc{Honest} \cite{nozza2021honest}, and \textsc{StereoSet} \cite{nadeem2021stereoset}. Racial biases are addressed in datasets such as \textsc{MRHC} \cite{bourgeade2023multilingual} and \textsc{DETESTS-Dis} \cite{schmeisser2024overviewdetests}, while biases against immigrants appear in \textsc{FB-Stereotypes} \cite{bosco2023detecting}, \textsc{StereoHoax} \cite{schmeisser2024stereohoax}, and \textsc{HaSpeeDe2} \cite{sanguinetti2020haspeede}. The inclusion of datasets dedicated to the LGBTQIA+ community, such as \textsc{Queereotypes} \cite{cignarella2024queereotypes}, demonstrates the growing awareness of biases affecting marginalized social groups beyond traditional categories like gender and race.
\medskip

\noindent \textit{\textbf{$\bullet$ Language and geographical focus.}} We observe that {6 out of the 15} datasets contain only English content and {3 out of the 15} contain content on more than one language at the same time. 
The overview suggests a strong emphasis on European languages, with English, Italian, Spanish, French and Portuguese appearing frequently. This suggests that much of the research in this dataset focuses on linguistic contexts that are well-resourced and commonly studied in NLP. The dominance of English often with regional specifications, such as English (India) and English (U.S.), indicates a consideration of geographical and cultural variations in language use.

Some datasets incorporate multiple languages, with mentions of gender-inflected languages like Italian, Romanian, and Spanish, suggesting a focus on linguistic structures that encode genders. Additionally, there are references to Hebrew (and very few other non-European languages). One dataset, SeeGULL \cite{jha2023seegull}, explicitly states that it includes stereotypes from 178 countries across multiple geopolitical regions, reflecting a more global perspective compared to other datasets that primarily focus on Western languages.
\medskip

\noindent \textit{\textbf{$\bullet$ Size.}}  
We observe that the datasets we encountered in our literature review vary a lot in size. 
{2 out of 15 datasets contain less than 1,000 sentences, while 10 out of them contain a few thousand annotated instances, such as the \textsc{FB-stereotypes} with 2,990 instances}. {One dataset} \citet{fraser2022extracting}{, comprises of 300,000 tweets.} This discrepancy reflects differences in data collection strategies, annotation efforts, and the computational resources available to different research groups.  

Larger datasets, such as \textsc{HaSpeeDe2} and \textsc{SeeGULL}, aim for broad coverage and diverse representation, often leveraging automated collection and annotation techniques. However, they may suffer from issues related to annotation quality and data noise. In contrast, smaller datasets, often manually curated, provide precise and high-quality annotations but may lack the scalability needed for robust model generalization. This variability in dataset size not only affects model training and evaluation but also highlights the challenges of balancing dataset breadth.
\medskip

\noindent \textit{\textbf{$\bullet$ Annotation.}} The literature reveals a diverse range of annotation approaches, reflecting different levels of expertise and methodology across the resources. Some datasets were annotated by in-house experts or researchers, ensuring a higher degree of linguistic or domain-specific knowledge in the labeling process. For example, one dataset was annotated by a mix of computational linguists, social psychologists, and linguistics students, which suggests a multidisciplinary approach to ensuring both technical and social science perspectives in the data labeling.

{There are 4 out of 15} datasets which relied on crowdsourcing platforms such as \textit{Amazon Mechanical Turk}, \textit{Figure8}, \textit{Appen}, and \textit{Prolific}, where workers were selected based on specific criteria, such as geographic location and task approval rates. This approach enables large-scale data annotation but may introduce quality control challenges, as crowdworkers typically lack specialized expertise. Additionally, some datasets were annotated using a semi-supervised approach, which combines human annotation with automated methods, striking a balance between efficiency and human judgment.

Notably, there is one dataset with no explicit annotation or cases where annotation details are unclear or not always specified. One dataset creation involved 89 annotators from 16 different countries, demonstrating a commitment to linguistic and regional diversity \cite{jha2023seegull}.
Overall, the annotation strategies vary widely, from expert-driven methods ensuring quality to crowd-sourced efforts enabling scalability, highlighting the trade-offs between precision, cost, and efficiency in dataset creation.
\medskip

\noindent \textit{\textbf{$\bullet$ Agreement.}} Regarding the measurement of inter-annotator agreement (IAA), we encountered a wide range of approaches, mostly dependent on the size of the corpus, the type of data and on the annotator pool. 
Some datasets rigorously assessed annotator consistency, while others either do not mention it or explicitly state that IAA was not calculated. The presence of various statistical methods such as Fleiss’ Kappa, Cohen’s Kappa, and Krippendorff’s Alpha suggests that for many datasets, ensuring annotation reliability was a priority. Some datasets report detailed IAA scores across multiple annotation categories (e.g., hate speech, offensiveness, irony, stereotypes, stance), indicating a more granular evaluation of agreement levels.
However, a significant number of datasets either do not calculate agreement or do not provide sufficient details. Additionally, some datasets mention specific cases where lower agreement was observed, particularly for neutral contexts, implying that subjectivity in annotation can be a challenge, especially in nuanced linguistic phenomena.
\medskip

\noindent \textit{\textbf{$\bullet$ Availability.}}  
We encountered a mixed landscape of dataset accessibility, indicating that while some datasets are openly shared, others have restricted or unclear availability. {8 datasets out of 15} are explicitly marked as publicly accessible, often via direct access through a URL link, connecting to repositories such as GitHub, OSF or HuggingFace.
However, there are 3 cases where the dataset is not available (e.g. broken link, or empty repository) and 4 cases where it is shared only upon specific request to the authors, which can introduce barriers to accessibility and slow down research progress.

Overall, while a good number of datasets are openly available, there is still inconsistency in accessibility across studies. Encouraging standardized data-sharing policies and ensuring proper dataset hosting would help improve the reproducibility and usability of research in this field.

\subsection{Languages}
As previously noted in the discussion on datasets (Section \ref{challenge:datasets}), the overall trend in the literature review reveals a strong predominance of English, reaffirming its position as the most extensively studied language in computational linguistics research.
Many of the papers reviewed focus on English-only, often with regional specifications such as English (U.S.) or English (India), suggesting that despite some awareness of national linguistic variations, the primary emphasis remains on Anglophone discourse. This dominance reflects the centrality of English in NLP advancements, but it is also limiting the applicability of findings to non-English contexts.

In addition to English, several papers include research in Italian, Spanish, French, Portuguese, Norwegian, Romanian and German, indicating a moderate presence of European languages in the literature. However, there is a notable absence (or very limited representation) of low-resource languages, which means that many linguistic communities, especially those in Africa, South Asia, and Indigenous groups, remain underrepresented in stereotype and bias-related studies. While the dataset includes Tamil and code-mixed Tamil-English \cite{chakravarthi2022homophobia}, such instances remain isolated rather than part of a broader effort to engage with underrepresented linguistic groups.

As far as national perspectives are concerned, relatively few papers seem to explicitly account for cross-cultural differences, though some studies attempt to analyze stereotypes across multiple geopolitical regions \cite{jha2023seegull}. However, in most cases, {with the exception of a recent contribution}\footnote{{This paper was published after our screening process and was therefore not included in the literature tables.}} by \citet{mitchell-etal-2025-shades}, multilingual research remains confined to European languages, and national perspectives outside of Western frameworks are rarely explored in any depth.

\subsection{Generalizability and Staticity}
Many models are trained on specific datasets that reflect narrow cultural perspectives, limiting their ability to transfer effectively to diverse linguistic environments \cite{jha2023seegull, ranaldi2024trip}. This raises concerns about \textit{domain adaptation}, as biases embedded in training data may affect how models interpret stereotypes in new settings.  
Another critical issue is that most models rely on static datasets, which fail to capture the evolving nature of stereotypes. Stereotypical associations shift over time, yet NLP models often struggle with temporal bias, leading to outdated predictions \cite{cao2022theorygrounded, mendelsohn2020dehumanization}. Moreover, implicit stereotypes remain difficult to detect due to their context-dependent nature, highlighting the need for \textit{adaptive learning approaches} and multilingual fine-tuning.  

Finally, inconsistencies in annotation and stereotype categorization hinder generalization. The absence of standardized frameworks causes models to overfit to specific datasets rather than acquiring robust detection strategies.

\subsection{Evaluation}
One of the most persistent challenges in stereotype detection in NLP is the difficulty of establishing a universally applicable evaluation framework. While existing benchmarks and well-known metrics, such as F-score, Accuracy, Precision and Recall, provide a structured means of assessing performance, the fundamental issue lies in the subjectivity and cultural dependence of defining what constitutes a stereotype.
The lack of consensus on a standardized evaluation procedure stems from the tension between broad, generalizable approaches and those tailored to specific social or cultural contexts. Addressing bias at a high level often results in overly abstract criteria, whereas focusing on specific stereotypes, such as those targeting a particular marginalized group, is at risk of being too narrow. Similarly, multilingual evaluation efforts frequently fail to capture cultural and geographic nuances, while monolingual approaches remain constrained to particular linguistic communities. 

A major issue in current evaluation frameworks is the oversimplification of bias metrics. Many studies attempt to capture bias through numerical aggregation methods, where models are assigned scores that quantify their level of stereotype reinforcement or fairness. However, bias in language is highly context-dependent and multifaceted, making it problematic to reduce such complexities to single numerical values. \citet{cao2022theorygrounded}, for example, caution against using averaged human judgments to assess bias, as this approach often collapses diverse perspectives into a single statistic, obscuring how different social groups perceive and experience stereotypes. Without a more nuanced way to analyze bias in NLP models, current benchmarks risk misrepresenting real-world social biases and overlooking failing to capture the intersecting nature of identity-based stereotypes.

Beyond dataset biases, benchmarking frameworks also struggle with real-world applicability. Many stereotype detection models are evaluated using static datasets, which fail to capture the evolving nature of stereotypes, biases, and societal attitudes over time. For example, datasets collected several years ago may no longer reflect current linguistic trends or emerging biases, making them less effective for evaluating contemporary NLP systems. Additionally, benchmarks often focus on explicit stereotypes and overt biases, neglecting subtler, implicit biases that require more sophisticated detection techniques. Studies such as \citet{schmeisser2022criteria} {highlight the need for more comprehensive datasets that go beyond surface-level annotations. They emphasize the importance of capturing implicit stereotypes, multimodal biases (such as those embedded in images, hashtags, and URLs) and the evolving nature of language across platforms and time. The absence of such datasets limits the ability of stereotype detection models to generalize across contexts and modalities. Without access to rich, diverse, and temporally-aware data, models risk reinforcing existing biases or failing to detect nuanced forms of stereotyping. Addressing this gap is crucial for developing more robust and socially responsible multimodal systems.}

Another key limitation in benchmarking is the lack of consistency in how bias mitigation techniques are evaluated. 
Some studies propose debiasing methods that work well under specific conditions, but these approaches do not always generalize across different datasets or linguistic settings.
Without a unified evaluation framework, it remains unclear whether a given mitigation strategy truly reduces bias or simply reconfigures it in a way that is less detectable under the existing benchmarks. Moreover, some bias mitigation strategies may introduce new biases or degrade model performance on other NLP tasks, which current benchmarks often fail to assess.

\section{CONCLUSIONS AND FUTURE WORK}\label{sec:conclusion}

This survey has examined the landscape of Stereotype Detection in Natural Language Processing, reviewing over 50+ studies that span diverse methodologies, datasets, and theoretical perspectives. Our literature review has highlighted that, while there is broad agreement on the fundamental nature of stereotypes -- often referring to established psychological theories such as those by Fiske and Koch -- many studies use \textit{stereotype}, \textit{bias}, and \textit{prejudice} interchangeably, leading to conceptual ambiguity. This lack of clear boundaries complicates the operationalization of stereotype detection in computational settings, making classification and evaluation inconsistent across studies.

Despite this initial challenge, and the relative novelty of the research field, NLP researchers have made substantial progress in advancing studies on stereotypes, particularly over the past 5 years. Recent work has proposed alternative debiasing techniques, refined stereotype taxonomies, and improved annotation frameworks. 
Additionally, interdisciplinary efforts drawing on social psychology, linguistics, and ethics have played a crucial role in contextualizing stereotype detection within broader societal and cultural frameworks. Moreover, the growing emphasis on transparency and reproducibility from the NLP/*CL community\footnote{See for instance: \url{https://aclrollingreview.org/static/responsibleNLPresearch.pdf}.} has led to an increase in publicly available datasets, facilitating further research and model benchmarking. However, there is still a pressing need to move beyond traditional, predominantly English-centric datasets and benchmarks to create more representative and culturally aware NLP models.

\subsection{Research Opportunities}\label{sec:opportunities}
From the information gathered during the literature review, we believe that future research should prioritize the following directions:
\smallskip

\noindent \textbf{\textit{Clarifying Conceptual Boundaries.}} One of the most pressing challenges is the need for clearer distinctions between stereotypes, biases, and prejudices. Many NLP studies conflate these terms, making it difficult to standardize annotation guidelines and evaluate models effectively. Future research should establish more precise theoretical frameworks that can be consistently operationalized in computational settings, reducing conceptual fuzziness and ambiguity.
\smallskip

\noindent \textbf{\textit{Exploring Alternative Stereotype Models.}}
Current stereotype detection frameworks largely rely on predefined taxonomies, but alternative models may offer deeper insights into how stereotypes manifest in language. \citet{davani2023socialstereotypes} propose the Agent-Beliefs-Communion (ABC) model as a more structured way to analyze stereotypes, while \citet{neveol2022frenchcrows} suggest leveraging Social Frames for cross-cultural bias characterization. \citet{fraser2021understanding} recommend leveraging large-scale social media corpora to analyze the emergence and evolution of stereotypes in real-world discourse. Future research should investigate these alternative models to develop more robust stereotype detection systems.
\smallskip

\noindent \textbf{\textit{Broadening the Scope of Target Attributes.}} Current research predominantly focuses on gender and racial stereotypes, often neglecting other important dimensions, such as sexual orientation or identity, (dis)ability, and socio-economic status. Moreover, most studies treat gender as a binary construct, failing to include non-binary identities. Another significant gap is the lack of research on intersectionality \cite{crenshaw1989intersectionality}. Future work should broaden coverage of these underrepresented attributes to achieve  a more comprehensive understanding of stereotypes in NLP.
\smallskip

\noindent \textbf{\textit{Enhancing Context Awareness in Models.}} Stereotypes are highly context-dependent, yet many existing NLP models struggle to differentiate between stereotypes that vary across cultural and situational contexts. Addressing this challenge requires models that go beyond simple classification and incorporate context-aware detection strategies, particularly for implicit stereotypes. Future research should explore ways to integrate socio-cultural awareness into NLP systems, potentially through the use of multimodal and discourse-aware models.
\smallskip

\noindent \textbf{\textit{Improving Dataset Creation and Annotation.}} 
The quality and scope of datasets remain a critical issue. Some studies rely on small, highly specific datasets, while others use large-scale corpora with minimal human annotation. This inconsistency affects model generalizability and reliability. 
Future efforts should aim to develop scalable datasets that strike a balance between annotation precision and diversity, while also embracing participatory approaches that actively involve the most affected communities and ensure a more inclusive representation of diverse subjectivities (e.g. \citet{felkner2023winoqueer,queerinai2023queer}).

\smallskip

\noindent \textbf{\textit{Expanding Linguistic and Cultural Representation.}} The majority of stereotype detection research focuses on English and a handful of widely spoken languages, leaving low-resource languages and regional dialects underrepresented. Future studies should aim to develop multilingual datasets that account for linguistic diversity and cultural nuances.
\smallskip

\medskip

\noindent This concludes our literature review on stereotype detection in NLP. While significant progress has been made, challenges remain in conceptual clarity, dataset diversity, and model generalizability. The lack of standardized evaluation frameworks and reliance on static datasets further hinder robust stereotype detection.
This research field offers exciting opportunities. Expanding linguistic representation, refining context-aware models, and integrating interdisciplinary insights can enhance fairness and inclusivity in NLP. Investigating these areas will be crucial for developing socially responsible language technologies.
\section*{LIMITATIONS\label{sec:limitations}}\hypertarget{h:limitations}{}
{This survey sought to provide an entry point into the field of stereotype detection in NLP by mapping key research trends, methodologies, and challenges. However, several limitations must be acknowledged to provide transparency of our approach.

First, this is \textit{not} a systematic review. A truly exhaustive analysis would have required in-depth examination of all 139+ retrieved papers, which was beyond the scope of this work. Instead, we focused on the most cited and influential studies to provide a broad yet selective overview. While this approach highlights major contributions, it may have overlooked less visible but equally important work.
Second, our study did not attempt to resolve ethical or fairness-related debates. These issues are inherently complex and require interdisciplinary perspectives, including insights from policy, philosophy, and social sciences. Our goal was to document existing research efforts, not to audit their ethical standing or prescribe normative solutions.

Third, we did not propose direct interventions for mitigating stereotypes or biases in NLP and ML models. Rather, we presented a synthesis of past and current approaches to help frame future research directions. This survey is intended as a foundation for scholars entering the field, offering a structured understanding of what has been done and what remains to be explored.
Additionally, our analysis reflects a fixed point in time (the first half of 2025) based on a dataset most recently updated in early February 2025. Given the rapid evolution of this research area, some findings may soon become outdated. We encourage future updates and follow-up studies to maintain relevance.

Finally, we acknowledge a positionality limitation: all authors come from the field of Computational Linguistics and NLP. While this background shapes our perspective, we recognize the importance of contributions from adjacent disciplines such as social psychology, sociology, and computational social science. Their insights are essential for a more comprehensive understanding of stereotypes in language technologies.
Despite these limitations, we believe this survey provides a timely and useful overview of the field, helping researchers navigate a rapidly growing area of study.}

\begin{acks}
{This work is funded by the European Union’s Horizon 2020 research and innovation programme under the Marie Skłodowska-Curie Actions, Grant Agreement No. 101146287. Views and opinions expressed are, however, those of the authors only and do not necessarily reflect those of the European Union or the European Research Executive Agency (REA). Neither the European Union nor the granting authority can be held responsible for them.}
\end{acks}

\bibliographystyle{ACM-Reference-Format}
\bibliography{bibliography}

\newpage
\appendix
\section{Appendix\label{appendix}}
Implementation details of BERTopic used for the topic analysis in Section \ref{subsubsec:analyses-on120paers}.

\begin{mintedbox}{python}
from bertopic import BERTopic
from sklearn.feature_extraction.text import CountVectorizer

topic_model = BERTopic(
    language="english",
    calculate_probabilities=True,
    verbose=True,
    n_gram_range=(1, 2),
    min_topic_size=10,
    embedding_model="all-MiniLM-L6-v2"
)

topics, probs = topic_model.fit_transform(all_abstracts_cleaned)
\end{mintedbox}

\newpage
\section{Appendix\label{appendixB}}
Categories used in~\cite{ortega2021irostereo}: national majority groups, illness/health groups, age and role family groups, victims, political groups, ethnic/racial minorities, immigration/national minorities professional and class groups, sexual orientation groups, women, physical appearance groups, religious groups, style of life groups, non-normative behavior groups, man/male groups, minorities expressed in generic terms and white people.\\

\noindent Categories used in~\cite{nangia2020crows}, ~\cite{neveol2022frenchcrows} and~\cite{parrish2022bbq}: Race/skin color, gender/gender identity or expression, socio-economic status/occupation, nationality, religion, age, sexual orientation, physical appearance, and disability.

\end{document}